# Near-Optimal Discrete Optimization for Experimental Design: A Regret Minimization Approach [*]

Zeyuan Allen-Zhu[1], Yuanzhi Li[2], Aarti Singh[3], and Yining Wang[3]

[1]Microsoft Research Redmond
[2]Princeton University
[3]Carnegie Mellon University
Emails: `zeyuan@csail.mit.edu`, `yuanzhi@princeton.edu`,
`{aarti,yiningwa}@cs.cmu.edu`

November 14, 2017

## Abstract

The experimental design problem concerns the selection of $k$ points from a potentially large design pool of $p$-dimensional vectors, so as to maximize the statistical efficiency regressed on the selected $k$ design points. Statistical efficiency is measured by *optimality criteria*, including A(verage), D(eterminant), T(race), E(igen), V(ariance) and G-optimality. Except for the T-optimality, exact optimization is NP-hard.

We propose a polynomial-time regret minimization framework to achieve a $(1+\varepsilon)$ approximation with only $O(p/\varepsilon^2)$ design points, for all the optimality criteria above.

In contrast, to the best of our knowledge, before our work, no polynomial-time algorithm achieves $(1+\varepsilon)$ approximations for D/E/G-optimality, and the best poly-time algorithm achieving $(1+\varepsilon)$-approximation for A/V-optimality requires $k = \Omega(p^2/\varepsilon)$ design points.

# 1 Introduction

Let $x_1, \ldots, x_n \in \mathbb{R}^p$ be $p$-dimensional vectors and $f: \mathbb{S}_p^+ \to \mathbb{R}^+$ be a non-negative function defined over $\mathbb{S}_p^+$, the class of all $p$-dimensional positive definite matrices. We focus on the design of polynomial-time algorithms for approximately solving the following *discrete* optimization problem:

$$\min_{s \in \mathcal{S}_k} F(s) = \min_{s \in \mathcal{S}_k} f\left(\sum_{i=1}^n s_i \cdot x_i x_i^\top\right) \quad \text{where} \quad \mathcal{S}_k := \left\{s \in \{0,1\}^n, \sum_{i=1}^n s_i \leq k\right\}. \quad (1.1)$$

In other words, we wish to select a subset $S \subset [n]$ of cardinality at most $k$, so that its covariance matrix $\Sigma_S = \sum_{i \in S} x_i x_i^\top$ has the smallest function value $f(\Sigma_S)$. The main challenge of solving Problem (1.1) is the discrete constraint $s \in \{0,1\}^n$.

More generally, we also study the multiplicity-$b$ generalization of Problem (1.1), that is

$$\min_{s \in \mathcal{S}_{k,b}} F(s) = \min_s f\left(\sum_{i=1}^n s_i \cdot x_i x_i^\top\right) \quad \text{where} \quad \mathcal{S}_{k,b} := \left\{s_i \in \{0, 1, \ldots, b\}, \sum_{i=1}^n s_i \leq k\right\}, \quad (1.2)$$

for any integer value $b$. It is a simple exercise to see that Problem (1.2) reduces to Problem (1.1) by duplicating each design point $x_i$ exactly $b$ times.[1] Throughout this paper we assume without loss of generality that $b \geq 1$ and $bn \geq k$.

## 1.1 Motivations

Before introducing our main assumptions on $f(\cdot)$, we discuss how a wide range of problems in experimental design and active learning can be cast as special cases of Problem (1.1):

**Classical experimental design.** The *(classical) experimental design* problem concerns the selection of $k$ points from a potentially very large design pool $\{x_1, \ldots, x_n\}$ so as to maximize the *statistical efficiency* regressed on the selected $k$ design points.

For example, consider a clinical study application where $n$ is the number of patients; $p$ is the number of parameters (e.g., blood pressure, low-density lipoprotein, etc.) that are hypothesized to affect some disease; and $x_1, \ldots, x_n \in \mathbb{R}^p$ are the parameters for all the patients. Since determining whether or not a patient has a certain disease may be expensive or time-consuming, one wishes to select $k \ll n$ patients that are the most *statistically efficient* for establishing a regression model that connects experimental parameters to the disease.

This experimental design problem reduces to Problem (1.1), where the evaluation of statistical efficiency is reflected in the choice of the objective function $f$, known as the *optimality criterion* [38]. Popular choices of $f$ include

- A(verage)-optimality $f_A(\Sigma) = \text{tr}(\Sigma^{-1})/p$,
- D(eterminant)-optimality $f_D(\Sigma) = (\det \Sigma)^{-1/p}$,
- T(race)-optimality $f_T(\Sigma) = p/\text{tr}(\Sigma)$,
- E(igen)-optimality $f_E(\Sigma) = \|\Sigma^{-1}\|_2$,
- V(araince)-optimality $f_V(\Sigma) = \frac{1}{n}\text{tr}(X\Sigma^{-1}X^\top)$, and
- G-optimality $f_G(\Sigma) = \max \text{diag}(X\Sigma^{-1}X^\top)$.

---
[1] Very often, an algorithm to Problem (1.1) can be directly generalized to solve Problem (1.2) without blowing up the problem size. Our algorithm proposed in this paper is one such algorithm.

We refer the readers to [38] for a complete list and discussion of various optimality criteria used in the experimental design literature. Note that the optimality criteria in the above list are "normalized" (by multiplying or raising to the power of $1/p$) so that $f(t\Sigma) = t^{-1}f(\Sigma)$.

*Remark* 1.1. If noise is statistically independent when we perform the same experiment multiple times, then the experiment design problem reduces to Problem (1.2) with $b = k$, and we can choose to include a design point $x_i$ multiple times.

*Remark* 1.2. Exact optimizing Problem (1.1) in T-optimality is trivial, because to maximize $\text{tr}(\Sigma)$, by linearity, it suffices to pick the $k$ distinct indices $i \in [n]$ to maximize $\text{tr}(x_i x_i^\top) = \|x_i\|^2$. On the other hand, optimizing Problem (1.1) in other optimality criteria are highly non-trivial. For instance, for D/E-optimality, exact optimization is NP-hard [17] as well as $(1 + \varepsilon)$ approximation without additional assumptions [24, 43]. A more detailed summary is given in Section 1.4.

**Bayesian experimental design.** In some applications of experimental design, a *prior* is imposed on the regression model [18]. In cases where a Gaussian prior $\mathcal{N}(0, \lambda)$ is imposed and the homogeneous noise is sub-Gaussian with parameter $\sigma^2$, the *Bayesian experimental design* problem reduces to Problem (1.1) or (1.2) with objective $f_{\lambda,\sigma}(\Sigma) = f(\lambda/\sigma^2 \cdot I_{p \times p} + \Sigma)$. Here, $f(\cdot)$ is any optimality criterion we have introduced in the classical experimental design setting.

Such objectives $f_{\lambda,\sigma}$ are also known as *Bayesian alphabetical optimality* [18] and are useful when the design budget $k$ is close to the number of parameters $p$, or when classical experiment design yields ill-conditioned solutions.

**Active learning.** Pool-based active learning plays an important role in label-efficient machine learning research [5], where a large pool of *unlabeled* data points $(x_1, \ldots, x_n)$ are available and the learning algorithm needs to select $k \ll n$ points in the pool to request (possibly human) labels.

One popular model of active learning is *active linear regression*, where real-valued labels $y_i$ are modeled as noisy realizations of inner product $\langle x_i, \beta \rangle$ with an unknown regression model $\beta$. This problem reduces to Problem (1.1), where the objective $f$ can be any optimality criterion in classical experimental design setting. Note that Problem (1.2) may not make sense for active learning, because each data point $x_i$ in the pool can usually be selected at most once. This constraint is mostly relevant in machine learning applications like image classification, where labels are unlikely to change if the same image is queried for more than once.

**Optimal subsampling in graph signal processing.** Graph signal processing has recently attracted significant attention in the statistical signal processing society [19, 21, 22, 23]. Suppose $G = (V, E, W)$ is an edge-weighted graph with $|V| = n$, and $L = D - A \in \mathbb{R}^{n \times n}$ is the graph Laplacian of $G$, where $D$ a diagonal degree matrix and $A$ is the adjacent matrix. Let $y \in \mathbb{R}^n$ be a noisy realization of an unknown vertex signal to be recovered.

In this setting, the graph signal is often assumed to have "weak high-frequency components", meaning that $y = U\theta + \varepsilon$, where $U \in \mathbb{R}^{n \times p}$ is the top-$p$ eigen-space of $L$, and $\varepsilon \in \mathbb{R}^n$ is the white noise. Ordinary least square regression can be applied to recover the "low-frequency signal" component $\theta$. In practice, however, it is often expensive to obtain the noisy signal $y$ on the entire graph $G$, as real-world graphs can have a huge number of nodes. It is thus desirable to *subsample* $k \ll n$ nodes in $G$, and approximately recover $\theta$ based on the limited measurements. Optimal subsampling therefore reduces to (classical or Bayesian) experiment design mentioned above.

## 1.2 Notations, Assumptions and Our Main Result

We write $A \succeq B$ if matrix $A - B$ is positive semi-definite (PSD), meaning that $x^\top(A - B)x \geq 0$ for all vectors $x$, and $A \succ B$ if $A - B$ is positive definite, meaning that $x^\top(A - B)x > 0$ for all non-zero vectors $x$. The matrix inner product $\langle A, B \rangle = \sum_{i,j} A_{ij} B_{ij} = \text{tr}(A^\top B)$. For symmetric matrices, we use $\lambda_{\max}(\cdot)$ and $\lambda_{\min}(\cdot)$ to denote their maximum and minimum eigenvalues. We use $\|A\|_{\text{op}} = \sqrt{\lambda_{\max}(A^\top A)}$ to denote the spectral norm of an arbitrary matrix $A$.

In this paper we make the following regularity assumptions on the objective $f : \mathbb{S}_p^+ \to \mathbb{R}^+$:

(A1) *Monotonicity*: for any $A, B \in \mathbb{S}_p^+$ with $A \preceq B$, it holds that $f(A) \geq f(B)$;

(A2) *Reciprocal sub-linearity*: for any $A \in \mathbb{S}_p^+$ and $t \in (0, 1)$, it holds that $f(tA) \leq t^{-1} f(A)$;

(A3) *Polynomial-time approximability of continuous relaxation*: for any fixed $\delta \in (0, 1)$, the *continuous relaxation* of Eq. (1.2) defined as

$$\min_{s \in \mathcal{C}_{k,b}} F(s) = \min_{s \in \mathcal{C}_{k,b}} f\left(\sum_{i=1}^n s_i \cdot x_i x_i^\top\right) \quad \text{where} \quad \mathcal{C}_{k,b} := \{s \in [0,b]^d : \sum_{i=1}^n s_i \leq k\}. \quad (1.3)$$

can be solved with $(1+\delta)$-relative error (i.e., $F(\widehat{s}) \leq (1+\delta) \min_{s \in \mathcal{C}_{k,b}} F(s)$) by a polynomial-time algorithm.

Assumption (A1) is natural in experimental design, because a design with sample covariance $A$ can never achieve superior statistical efficiency than a design with $B$ if $A \preceq B$.

Assumption (A2) captures the linearity in *classical* experiment design, where a design with sample covariance $tA$ is exactly $t$-times as efficient as a design with covariance $A$. Assumption (A2) also holds in Bayesian experiment design: the additional smoothing term $\lambda/\sigma \cdot I_{p \times p}$ ensures $f(tA) < t^{-1} f(A)$ for all $0 < t < 1$.

**Fact 1.3.** *(A1) and (A2) are satisfied for A/D/T/E/V/G optimality as well as their Bayesian counterparts.*

Assumption (A3) is motivated by the observation that, for most popular optimality criteria the function $f$ itself or one of its "surrogates" is *convex*, meaning that its *continuous relaxation* can is efficiently solvable. In Section 3 we validate (A3) for all A/D/T/E/V/G optimality criteria by considering an entropic mirror descent solver [11], which enjoys rigorous theoretical guarantees and also runs fast in practice.

Below is our main result:

**Theorem 1.4.** *Suppose $\varepsilon \in (0, 1/3]$, $n \geq k \geq 5p/\varepsilon^2$, $b \in \{1, 2, \ldots, k\}$, $f : \mathbb{S}_p^+ \to \mathbb{R}$ satisfies assumptions (A1)-(A3), and $\min_{s \in \mathcal{S}_{k,b}} F(s) < +\infty$. Then, there exist a polynomial-time algorithm that outputs*

$$\widehat{s} \in \mathcal{S}_{k,b} \quad \text{satisfying} \quad F(\widehat{s}) \leq (1 + 8\varepsilon) \cdot \min_{s \in \mathcal{S}_{k,b}} F(s) \ .$$

(To prove the simplest proofs, we have not tried to improve the constants in this statement.)

Note that Theorem 1.4 makes assumption $k \geq \Omega(p/\varepsilon^2)$ in order to achieve $(1+\varepsilon)$-approximation. In contrast, without any additional assumption, at least for D/E-optimality, it is NP-hard to achieve any $(1 + \varepsilon)$-approximation. [24, 43]

*Remark* 1.5. Our assumption $k \geq \Omega(p/\varepsilon^2)$ is quite mild. A lower bound $k \geq p$ is needed because if $k < p$ then no $s \in \mathcal{S}_k$ can produce a covariance matrix $\sum_{i=1}^n s_i \cdot x_i x_i^\top$ that is full rank. We conjecture the $\varepsilon^{-2}$ dependency is necessary at least for E-optimality (and thus also for general objectives $f$), motivated by negative results in spectral graph theory, see Section 5. In our experiments, we shall demonstrate that our algorithm even supports values such as $k = 1.2p$.

*Remark* 1.6. While preparing the journal version of this paper, we were informed of an independent work of [41], where the authors obtained a $(1+\varepsilon)$ approximation for the D-optimality criterion with $k = \Omega\big(\frac{p}{\varepsilon} + \frac{\log \varepsilon^{-1}}{\varepsilon^2}\big)$. Their algorithm is unlikely to be extendable to general objectives $f(\cdot)$ studied in this paper.

### 1.3 Overview of Techniques

The main technical challenge is to round the fractional solution (from continuous optimization) into an integral one. For instance, given $\pi \in [0,1]^n$ with $\|\pi\|_1 \leq k$ which is a fractional solution to Problem (1.1), we want to round it to a vector $s \in \{0,1\}^n$ with sparsity at most $k$, so that the covariance matrix $\sum_{i\in[n]} s_i \cdot x_i x_i^\top$ performs $(1+\varepsilon)$ close to the fractional covariance matrix $\sum_{i\in[n]} \pi_i \cdot x_i x_i^\top$, for *any* function $f(\cdot)$.

Using assumptions (A1) and (A2), we reduce this task to showing $\sum_{i\in[n]} s_i \cdot x_i x_i^\top \succeq (1-\varepsilon)\sum_{i\in[n]} \pi_i \cdot x_i x_i^\top$ without any explicit information of $f(\cdot)$. Then, after rotating the space, we rewrite this problem as "lower bounding the minimum eigenvalue" of $\sum_{i\in[n]} s_i \cdot x_i x_i^\top$.

Our main contribution is to reduce this eigenvalue lower-bounding process to regret minimization in online learning. Very informally, starting from an arbitrary set $S_0$ of cardinality $k$, we consider the following iterative process. In each iteration $t$,

- the player selects a density matrix $A_t \in \Delta_{p \times p} := \{A \succeq 0 : \mathrm{tr}(A) = 1\}$ which "allegedly minimizes" $\langle A_t, \sum_{i \in S_t} x_i x_i^\top \rangle$; and

- the adversity independently selects a swap $S_t = S_{t-1}\setminus\{j\}\cup\{i\}$ so as to "allegedly maximize" $\langle A_t, \sum_{i \in S_t} x_i x_i^\top \rangle$.

We have adopted the modern $\ell_{1/2}$-regularized strategy [4] for the player, and proved a new regret theorem. We have also introduced a strategy for the adversary so that he/she can always find a good swapping pair $(i,j)$ using a linear scan to the vectors $x_1, \ldots, x_n$.

Finally, if we run $T$ rounds, the "value" of the game is close to
$$\min_{A \in \Delta_{p \times p}}\{\langle A, \sum_{i \in S_T} x_i x_i^\top \rangle\} = \lambda_{\min}(\sum_{i \in S_T} x_i x_i^\top) \ .$$

Therefore, a regret theorem necessarily gives a lower bound to the value of the game, which then gives a lower bound to the minimum eigenvalue of $\sum_{i \in S_T} x_i x_i^\top$. In particular, we achieve the desired lower bound as long as $T = \Omega(k/\varepsilon)$ and $k \geq \Omega(p/\varepsilon^2)$.

### 1.4 Related Work

Experimental design is an old topic in statistics [18, 30, 38]. Computationally efficient experimental design algorithms (with provable guarantee) are, however, a less studied field.

Perhaps the most well-studied optimality criterion is D-optimality, whose negative logarithm (i.e., $\log \det \Sigma$) is *submodular*, a property that sometimes gives rises to $1 - 1/e$ approximation ratio using pipage rounding [1]. Unfortunately, $\log \det \Sigma$ can be negative and thus pipage rounding fails to directly solve (1.1) with D-optimality. In [14], Bouhtou et al. proposed to maximize a function $h(\Sigma) := \frac{1}{p}\text{tr}(\Sigma^q)$ for $q \in (0,1]$, and it satisfies $\lim_{q \to 0}(h(\Sigma))^{-1/q} = f_D(\Sigma)$. They showed that $h(\Sigma)$ is submodular and gave a $(1 - 1/e)$ approximation to $h(\Sigma)$ for every $q \in (0,1]$ using pipage rounding. This does not translate to any bounded approximation ratio for $f_D(\Sigma)$ because $(1-1/e)^{-1/q}$ is unbounded when $q$ approaches zero.

| paper | solves (1.1)? | deterministic? | constraints | criteria | appx. ratio |
|---|---|---|---|---|---|
| pipage rounding [14] | Yes | Yes | $k \geq p$ | D | unbounded[c] |
| Nikolov-Singh [37] | Yes | No | $k = p$ | D | $e$ |
| Avron-Boutsidis [8] | Yes | No | $k \geq p$ | A | $\frac{n-p+1}{k-p+1}$ |
| Wang-Yu-Singh [46] | Yes | Yes | $k \geq \Omega(p^2/\varepsilon)$ | A,V | $1+\varepsilon$ |
| our pre. version [3] | Yes | Yes | $k > 2p$ | A,D,E,V,G | $O(1)$ |
| **this paper** | **Yes** | **Yes** | $k \geq \Omega(p/\varepsilon^2)$ | **A,D,E,V,G** | $1+\varepsilon$ |
| Wang-Yu-Singh [46] | No ($b = k$ only[b]) (oversampling[a]) | No | $k \geq \Omega((p\log p)/\varepsilon^2)$ | A,V | $1+\varepsilon$ |
| our pre. version [3] | No ($b = k$ only[b]) | Yes | $k \geq \Omega(p/\varepsilon^2)$ | A,D,E,V,G | $1+\varepsilon$ |
| our pre. version [3] | No (oversampling[a]) | Yes | $k \geq \Omega(p/\varepsilon^2)$ | A,D,E,V,G | $1+\varepsilon$ |

Table 1: Comparison with existing polynomial-time algorithms on all popular optimality criteria. We ignore T-optimality because it is trivially solvable. An algorithm produces a subset $\widehat{S} \in \mathcal{S}_{k,b}$, and the approximation ratio is defined as $f(X_{\widehat{S}}^\top X_{\widehat{S}})/\min_{S \in \mathcal{S}_{r,b}} f(X_S^\top X_S)$.
[a] they output subsets $S$ of cardinality $O(k)$ as opposed to $k$, so do not solve Problem (1.1).
[b] they only solve the special case $b = k$ for Problem (1.2), and thus not Problem (1.1).
[c] see our related work Section 1.4.

Summa et al. [43] gave a polynomial-time algorithm for a related maximum volume simplex (MVS) problem in computational geometry with an $O(\log p)$ approximation ratio, which was later improved to $O(1)$ by [36, 37]. Their results imply an $e$ approximation ratio for Eq. (1.1) in the special case of $k = p$. On the other hand, Summa et al. [43] showed that there exists a constant $c > 1$ such that polynomial-time $c$-approximation of the D-optimality is impossible for the $p = k$ case, unless $\text{P} = \text{NP}$. Therefore, additional assumptions on $k$ are necessary for the $(1+\varepsilon)$-approximation regime we consider in this paper.

Another well-studied optimality criterion is the A/V-optimality, which is not submodular and hence pipage rounding no longer relevant. The papers [13, 19] considered an alternative "approximate supermodular" formulation and derived a greedy algorithm with an $O(1)$ approximation ratio for the A/V-optimality. Their results, however, only apply to Bayesian experimental design settings and require the number of samples ($k$) to be lower bounded by a quantity that depends on the condition number of the original design, which might be unbounded.

For the A-optimality criterion, Avron and Boutsidis [8] proposed a greedy algorithm with an approximation ratio $O(n/k)$ with respect to $f(X^\top X)$. This ratio is tight for their algorithm in the worst case.[2] Li et al. [32] further accelerated this greedy algorithm, and achieved similar approximation guarantees.

---
[2] In the worst case, even the exact minimum $\min_{|S| \leq k} f(X_S^\top X_S)$ can be indeed $O(n/k)$ times larger than $f(X^\top X)$ [8]. This worst-case scenario may not always happen, but to the best our knowledge, their proof is tight in this worst case.

Perhaps closest to this work, for A/V-optimality, Wang et al. [46] considered a variant of this greedy algorithm of [8], and proved an approximation ratio quadratic in design dimension $p$ and independent of pool size $n$. This result can also be turned into an $1 + \varepsilon$ approximation but requiring $k \geq \Omega(p^2/\varepsilon)$.

For the same A/V-optimality criteria, Wang et al. [46] derived another algorithm based on effective-resistance sampling [42], and attains a $(1+\varepsilon)$ *pseudo*-approximation ratio if $k = \Omega(p \log p/\varepsilon^2)$. Specifically, their output set $S$:

- is of cardinality $O(k)$ rather than $k$, and
- is a multi-set rather than a set for Problem (1.1).

This result is based on repeatedly "re-weighting" design points, so cannot output a set rather than a multi-set. Nevertheless, it is still an improvement to previous sampling-based methods [20, 26, 46], which also achieve $(1 + \varepsilon)$ approximations but require the subset size $k$ to be much larger than the condition number of $X$.

Our preliminary version of this paper [3] obtained three weaker variants of Theorem 1.4, all being outperformed by our new Theorem 1.4. In particular, the *true* approximation ratio of [3] is only $O(1)$ but we have $1 + \varepsilon$ in this paper. Although their algorithm is also based on regret minimization, it has naively applied the old regret theorem of [4] and thus given weaker results. In this paper, we have developed *a new regret theorem and a new swapping algorithm* which have not appeared in [3] or any other related work. They together enable us to obtain the exact Theorem 1.4.

A less relevant topic is low-rank matrix column subset selection and CUR approximation, which seeks column subset $C$ and row subset $R$ such that $\|X - CC^\dagger X\|_F$ and/or $\|X - CUR\|_F$ are minimized [15, 27, 28, 44, 45]. These problems are unsupervised in nature and do not in general correspond to statistical properties under supervised regression settings.

## 2 Rounding via Regret Minimization

In this section we prove our main rounding theorem:

**Theorem 2.1** (rounding). *Suppose $\varepsilon \in (0, 1/3]$, $n \geq k \geq 5p/\varepsilon^2$, $b \in \{1, 2, \ldots, k\}$, and $f : \mathbb{S}_p^+ \to \mathbb{R}$ satisfies assumptions (A1) and (A2). Let $\pi \in \mathcal{C}_{k,b}$ by any fractional solution so that $F(\pi) < +\infty$. Then, in time complexity $\widetilde{O}(np^2)$ we can round $\pi$ to an integral solution*

$$\widehat{s} \in \mathcal{S}_{k,b} \quad \text{satisfying} \quad F(\widehat{s}) \leq (1 + 6\varepsilon)F(\pi) \ .$$

We immediately note that Theorem 2.1 plus assumption (A3) yield Theorem 1.4. This is because, one can first run continuous optimization (see Section 3) to find a point $\pi \in \mathcal{C}_{k,b}$ satisfying $F(\pi) \leq (1 + \varepsilon/2) \min_{\pi^* \in \mathcal{C}_{k,b}} F(\pi^*) \leq (1 + \varepsilon/2) \min_{s \in \mathcal{S}_{k,b}} F(s)$, and then apply Theorem 2.1 on $\pi$. Since $\varepsilon \in (0, 1/3]$, we have $(1 + 6\varepsilon)(1 + \varepsilon/2) \leq (1 + 8\varepsilon)$ and this gives the desired approximation ratio in Theorem 1.4.

For notational simplicity, we only consider the case of $b = 1$. The general case of $b \geq 1$ can be handled by simply repeating each $x_i$ exactly $b$ times, and our proposed algorithm will have the same total computational complexity (without blowing up by a factor of $b$). We denote by $\mathcal{S}_k = \mathcal{S}_{k,1}$ and $\mathcal{C}_k = \mathcal{C}_{k,1}$.

Without loss of generality, throughout this section we assume that $\widehat{\Sigma} = \sum_{i=1}^n \pi_i x_i x_i^\top$ is invert-ible. If $\widehat{\Sigma}$ is singular instead, one can remove all $x_i$ that does not belong to the span of $\widehat{\Sigma}$ (such $x_i$ must be associated with $\pi_i = 0$) and project the rest of $x_i$ onto a $\text{rank}(\Sigma)$-dimensional linear space. The covariance $\widehat{\Sigma}$ would then be invertible in the projected low-dimensional space.

## 2.1 The Whitening Trick

Due to the monotonicity (A1) and reciprocal sub-linearity (A2) properties of $f$, we claim that it suffices to find an integral solution $\widehat{s} \in \mathcal{S}_{k,b}$ that satisfies

$$\left(\sum_{i=1}^n \widehat{s}_i x_i x_i^\top\right) \succeq \tau \cdot \left(\sum_{i=1}^n \pi_i x_i x_i^\top\right) \tag{2.1}$$

for some constant $\tau = 1 - O(\varepsilon) > 0$. This is so because (A1) and (A2) tell us that Eq. (2.1) implies

$$f\left(\sum_{i=1}^n \widehat{s}_i x_i x_i^\top\right) \leq f\left(\tau \sum_{i=1}^n \pi_i x_i x_i^\top\right) \leq \tau^{-1} f\left(\sum_{i=1}^n \pi_i x_i x_i^\top\right) \ .$$

In the rest of this section, we let $\Pi = \text{diag}(\pi)$ and $S = \text{diag}(\widehat{s})$ be $n \times n$ diagonal matrices, and let $X = [x_1^\top, \ldots, x_n^\top]^\top \in \mathbb{R}^{n \times p}$.

Consider now a linear transform $x_i \mapsto (X\Pi X^\top)^{-1/2} x_i =: \widetilde{x}_i$. It is easy to verify that $\sum_{i=1}^n \pi_i \widetilde{x}_i \widetilde{x}_i^\top = I$. Such transform is usually referred to as *whitening*, because the sample covariance of the transformed data is the identity matrix. If we define $W = \sum_{i=1}^n \widehat{s}_i \widetilde{x}_i \widetilde{x}_i^\top$, then

**Proposition 2.2.** $W \succeq \tau I$ if and only if $(\sum_{i=1}^n \widehat{s}_i x_i x_i^\top) \succeq \tau(\sum_{i=1}^n \pi_i x_i x_i^\top)$.

*Proof.* The proposition holds because $W \succeq \tau I$ if and only if $(X\Pi X^\top)^{1/2} W (X\Pi X^\top)^{1/2} \succeq \tau X\Pi X^\top$, and that $(X\Pi X^\top)^{1/2} W (X\Pi X^\top)^{1/2} = XSX^\top$. $\square$

Proposition 2.2 means that, without loss of generality, we may assume $\sum_{i=1}^n \pi_i x_i x_i^\top = X\Pi X^\top = I$. The question of proving $W = XSX^\top \succeq \tau I$ is then reduced to lower bounding the minimum eigenvalue of $W = \sum_{i=1}^n \widehat{s}_i \widetilde{x}_i \widetilde{x}_i^\top$. We summarize this new problem as follows:

**The minimum eigenvalue problem..** Suppose $\pi \in \mathcal{C}_k = \{\pi \in [0,1]^n \colon \sum_{i=1}^n \pi_i \leq k\}$ and it satisfies $\sum_{i=1}^n \pi_i x_i x_i^\top = I_{p \times p}$. Then, find $\widehat{s} \in \mathcal{S}_k = \{s \in \{0,1\}^n \colon \sum_{i=1}^n s_i \leq k\}$ such that

$$\lambda_{\min}\left(\sum_{i=1}^n \widehat{s}_i x_i x_i^\top\right) \geq (1 - 3\varepsilon) \cdot I. \tag{2.2}$$

With Eq. (2.2) being true, we conclude that $F(\widehat{s}) \leq (1 + 6\varepsilon) F(\omega^*)$ by applying the reciprocal sub-linearity of $f$ (A3) and the fact that $\varepsilon \in (0, 1/3]$.

## 2.2 Minimum Eigenvalue via Regret Minimization

We review the concept of regret minimization in the linear (matrix) optimization setting. Let $\Delta_{p \times p} = \{A \in \mathbb{R}^{p \times p} : A \succeq 0, \text{tr}(A) = 1\}$ be an *action space* that consists of PSD matrices of unit trace (a.k.a. density matrices).

Consider an iterative game for $T$ iterations. At iteration $t$, the player chooses an action $A_t \in \Delta_{p \times p}$; afterwards, a loss matrix $F_t$ is revealed and the player suffers loss $\langle F_t, A_t \rangle = \text{tr}(F_t^\top A_t)$. The goal of the player is to minimize his/her *regret*:

$$\text{Regret}(\{A_t\}_{t=0}^{T-1}) := \sum_{t=0}^{T-1} \langle F_t, A_t \rangle - \min_{U \in \Delta_{p \times p}} \sum_{t=0}^{T-1} \langle F_t, U \rangle, \tag{2.3}$$

which is the "excess loss" of $\{A_t\}_{t=0}^{T-1}$ compared to the single optimal action $U \in \Delta_{p \times p}$ in "hindsight" (knowing all the loss matrices $\{F_t\}_{t=0}^{T-1}$).

We immediately observe that the second term $\left(\min_{U \in \Delta_{p \times p}} \sum_{t=0}^{T-1} \langle F_t, U \rangle\right)$ in (2.3) is precisely the minimum eigenvalue of $\sum_{t=0}^{T-1} F_t$. Hence, the task of lower bounding $\lambda_{\min}(\sum_{t=0}^{T-1} F_t)$ can be reduced to upper bounding the regret in Eq. (2.3).

**Follow the regularized leader.** A popular strategy to minimize regret for the player is *Follow-The-Regularized-Leader (FTRL)*, also known equivalent to *Mirror Descent (MD)* [34]. It specifies strategy $A_t$ for player at each round $t = 0, 1, \ldots, T-1$ as follows:

$$A_t = \arg\min_{A \in \Delta_{p \times p}} \{\Delta_\psi(A_{t-1}, A) + \alpha \langle F_{t-1}, A \rangle\} \ . \tag{2.4}$$

Above, $\alpha > 0$ is the learning rate, $\psi : \mathbb{R}^{p \times p} \to \mathbb{R}$ is some differentiable *regularizer* function, and $\Delta_\psi(A, B) = \psi(B) - \psi(A) - \langle \nabla \psi(A), B - A \rangle$ is the so-called Bregman divergence function associated with $\psi$.

Perhaps the most famous choice of $\psi$ is the matrix entropy $\psi(A) = \langle A, \log A - I \rangle$, and the resulting FTRL strategy is $A_t = \exp\left(c_t I - \alpha \sum_{\ell=0}^{t-1} F_\ell\right)$ where $c_t$ is the normalization constant that ensures $\text{tr}(A_t) = 1$. This is often referred to as the matrix multiplicative weight updates [6].

**Our $\ell_{1/2}$ strategy.** In this paper, to achieve better regret, we adopt the less famous $\ell_{1/2}$-regularizer $\psi(A) = -2\text{tr}(A^{1/2})$ introduced in [4], and call the resulting FTRL strategy the $\ell_{1/2}$ strategy.

*Remark* 2.3. The vector version of the $\ell_{1/2}$ strategy was first introduced in [7] to obtain optimum regret for combinatorial prediction games. The matrix generalization of this $\ell_{1/2}$ strategy (or more generally the $\ell_{1-1/q}$ strategy for $q \geq 2$) is non-trivial, and leads to better algorithms for graph sparsification [4] and online eigenvector [2].

The following claim gives a closed form representation of the $\ell_{1/2}$ strategy. Its proof is by careful manipulations of the definition of $A_t$, and has implicitly appeared in [4]. We include it in the appendix for completeness' sake.

**Claim 2.4** (closed form $\ell_{1/2}$ strategy). *Assume without loss of generality that $A_0 = (c_0 I + \alpha Z_0)^{-2}$ for some $c_0 \in \mathbb{R}$ and symmetric matrix $Z_0$ such that $c_0 I + \alpha Z_0 \succ 0$. Then,*

$$A_t = \left(c_t I + \alpha Z_0 + \alpha \sum_{\ell=0}^{t-1} F_\ell\right)^{-2}, \quad t = 1, 2, \ldots, \tag{2.5}$$

*where $c_t \in \mathbb{R}$ is the unique constant so that $c_t I + \alpha Z_0 + \alpha \sum_{\ell=0}^{t-1} F_\ell \succ 0$ and $\text{tr}(A_t) = 1$.*

At a high level, if $Z_0 = 0$ were the zero matrix, then $A_0 = \frac{I}{\sqrt{p}}$ would be a multiple of identity. This corresponds to the standard way to initialize the player's strategy in online learning, and was used in [4]. In this paper, we need this more general $Z_0$ to support the swapping algorithm in the next subsection.

If each loss matrix $F_t$ can be rank-2 decomposed as $F_t = u_t u_t^\top - v_t v_t^\top$, then we show in Section 2.4 the following lemma which upper bounds the total regret of the $\ell_{1/2}$ strategy:

**Lemma 2.5** (main regret lemma). *Suppose $F_t = u_t u_t^\top - v_t v_t^\top$ for vectors $u_t, v_t \in \mathbb{R}^p$, and $A_0, \ldots, A_{T-1} \in \Delta_{p \times p}$ are defined according to the $\ell_{1/2}$ strategy with some learning rate $\alpha > 0$.*

Then, as long as $\alpha \langle A_t^{1/2}, v_t v_t^\top \rangle < 1/2$ for all $t$, we have for any $U \in \Delta_{p \times p}$,

$$-\sum_{t=0}^{T-1} \langle F_t, U \rangle \leq \sum_{t=0}^{T-1} \left( -\frac{\langle A_t, u_t u_t^\top \rangle}{1 + 2\alpha \langle A_t^{1/2}, u_t u_t^\top \rangle} + \frac{\langle A_t, v_t v_t^\top \rangle}{1 - 2\alpha \langle A_t^{1/2}, v_t v_t^\top \rangle} \right) + \frac{\Delta_\psi(A_0, U)}{\alpha}. \quad (2.6)$$

*Remark* 2.6. To see why Lemma 2.5 is a bound on regret (2.3), we rearrange the two sides:

$$\sum_{t=0}^{T-1} \langle F_t, A_t - U \rangle \leq 2\alpha \sum_{t=0}^{T-1} \left( \frac{\langle A_t, u_t u_t^\top \rangle \cdot \langle A_t^{1/2}, u_t u_t^\top \rangle}{1 + 2\alpha \langle A_t^{1/2}, u_t u_t^\top \rangle} + \frac{\langle A_t, v_t v_t^\top \rangle \cdot \langle A_t^{1/2}, v_t v_t^\top \rangle}{1 - 2\alpha \langle A_t^{1/2}, v_t v_t^\top \rangle} \right) + \frac{\Delta_\psi(A_0, U)}{\alpha}.$$

The proof of Lemma 2.5 involves a non-classical regret analysis designed for the matrix $\ell_{1/2}$ strategy. It is based on the closed-form expressions in Eq. (2.5). Note that a variant of Lemma 2.5, but only for matrices $F_t = u_t u_t^\top$ (thus of rank 1) was originally presented in [4, Theorem 3.2]. The involvement of the extra $-v_t v_t^\top$ components is, however, a non-trivial extension and brings in extra technical difficulties. The complete proof is given in Section 2.4.

We also need the following lemma to bound the Bregman divergence term $\Delta_\psi(A_0, U)$:

**Lemma 2.7.** *Suppose $A_0 = (c_0 I + \alpha Z_0)^{-2}$ as in Claim 2.4, then for any $U \in \Delta_{p \times p}$:*

$$\Delta_\psi(A_0, U) \leq 2\sqrt{p} + \alpha \langle Z_0, U \rangle \ .$$

*Proof of Lemma 2.7.* By definition of $\Delta_\psi$, $\psi$ and $A_0$, we have

$$\Delta_\psi(A_0, U) = \langle A_0^{-1/2}, U \rangle + \text{tr}(A_0^{1/2}) - 2\text{tr}(U^{1/2}) \leq \langle c_0 I + \alpha Z_0, U \rangle + \sqrt{p},$$

where the last inequality holds because $\text{tr}(U^{1/2}) \geq 0$ and $\text{tr}(A_0^{1/2}) \leq \sqrt{p \cdot \text{tr}(A_0)} = \sqrt{p}$. Note also that $\langle I, U \rangle = \text{tr}(U) = 1$. Therefore,

$$\Delta_\psi(A_0, U) \leq \alpha \langle Z_0, U \rangle + c_0 + \sqrt{p}.$$

Because $\text{tr}(A_0) = 1$, the constant $c_0$ (if positive) must be upper bounded by $\sqrt{p}$ because otherwise $\text{tr}(A_0) \leq \text{tr}((c_0 I)^{-2}) = p \cdot c_0^{-2} < 1$. Therefore, it is proved that $\Delta_\psi(A_0, U) \leq \alpha \langle Z_0, U \rangle + 2\sqrt{p}$. □

## 2.3 Our Swapping Algorithm

Lemma 2.5 and Lemma 2.7 together give rise to an interesting "swapping" technique for solving Problem (2.2). Suppose we start with an arbitrary initial set $S_0 \subseteq [n]$ of cardinality $k$, where $\sum_{j \in S_0} x_j x_j^\top$ may or may not have its minimum eigenvalue being very small. Next, in each iteration $t \geq 0$, we select a pair of indices $i_t \in S_t$ and $j_t \notin S_t$, and make a "swap" by updating $S_{t+1} = S_t \cup \{j_t\} \setminus \{i_t\}$.

In order to estimate how the minimum eigenvalue changes after the swap, we define $Z_0 = \sum_{j \in S_0} x_j x_j^\top$ and $F_t = x_{j_t} x_{j_t}^\top - x_{i_t} x_{i_t}^\top$, and apply Lemma 2.5 and Lemma 2.7. They together imply an upper bound of the form

$$-\left\langle Z_0 + \sum_{t=0}^{T-1} F_t, U \right\rangle \leq \sum_{t=0}^{T-1} \left( -\frac{\langle A_t, x_{j_t} x_{j_t}^\top \rangle}{1 + 2\alpha \langle A_t^{1/2}, x_{j_t} x_{j_t}^\top \rangle} + \frac{\langle A_t, x_{i_t} x_{i_t}^\top \rangle}{1 - 2\alpha \langle A_t^{1/2}, x_{i_t} x_{i_t}^\top \rangle} \right) + \frac{2\sqrt{p}}{\alpha} \ , \quad (2.7)$$

for any $U \in \Delta_{p \times p}$. Since we can choose $U \in \Delta_{p \times p}$ so that the left hand side exactly equals $-\lambda_{\min}(Z_0 + \sum_{t=0}^{T-1} F_t) = -\lambda_{\min}(\sum_{j \in S_T} x_j x_j^\top)$, Eq. (2.7) gives a lower bound on the minimum eigenvalue of matrix $\left(\sum_{j \in S_T} x_j x_j^\top\right)$.

What remains next is to find $i_t$ and $j_t$ so that the right hand side of Eq. (2.7) is as small as possible. The following lemma shows that, unless the minimum eigenvalue of $\sum_{j \in S_t} x_j x_j^\top$ is already large, there exists a pair of good indices $(i_t, j_t)$:

**Lemma 2.8** (main averaging lemma). *For every $\varepsilon > 0$ and subset $S \subseteq [n]$ of cardinality $k$, suppose $\lambda_{\min}(\sum_{i \in S} x_i x_i^\top) \leq 1 - 3\varepsilon$ and $A = (cI + \alpha \sum_{i \in S} x_i x_i^\top)^{-2}$, where $c \in \mathbb{R}$ is the unique number such that $A \succeq 0$ and $\mathrm{tr}(A) = 1$. Then, the following statements are true:*

$$\nu := \min_{i \in S, 2\alpha \langle A^{1/2}, x_i x_i^\top \rangle < 1} \frac{\langle A, x_i x_i^\top \rangle}{1 - 2\alpha \langle A^{1/2}, x_i x_i^\top \rangle} \leq \frac{1 - \varepsilon}{k}; \tag{2.8}$$

$$\max_{j \in [n] \setminus S} \frac{\langle A, x_j x_j^\top \rangle}{1 + 2\alpha \langle A^{1/2}, x_j x_j^\top \rangle} \geq \nu + \frac{\varepsilon}{k}. \tag{2.9}$$

*Furthermore, if $\alpha = \sqrt{p}/\varepsilon$ and $k \geq 5p/\varepsilon^2$ for some $\varepsilon \in (0, 1/3]$, then there always exists $i \in S$ such that $2\alpha \langle A^{1/2}, x_i x_i^\top \rangle < 1$.*

In other words, Lemma 2.8 suggests that, as long as $\lambda_{\min}(\sum_{i \in S} x_i x_i^\top) \leq 1 - 3\varepsilon$, we can simply choose $i_t$ to be the index $i \in S_t$ which minimizes $\frac{\langle A_t, x_i x_i^\top \rangle}{1 - 2\alpha \langle A_t^{1/2}, x_i x_i^\top \rangle}$, and $j_t$ to be the index $j \notin S_t$ which maximizes $\frac{\langle A_t, x_j x_j^\top \rangle}{1 + 2\alpha \langle A_t^{1/2}, x_j x_j^\top \rangle}$. Eqs. (2.8) and (2.9) together imply that

$$\left(-\frac{\langle A_t, x_{j_t} x_{j_t}^\top \rangle}{1 + 2\alpha \langle A_t^{1/2}, x_{j_t} x_{j_t}^\top \rangle} + \frac{\langle A_t, x_{i_t} x_{i_t}^\top \rangle}{1 - 2\alpha \langle A_t^{1/2}, x_{i_t} x_{i_t}^\top \rangle}\right) \leq -\frac{\varepsilon}{k}. \tag{2.10}$$

In sum, either there exists some index $t = 0, 1, \ldots, T-1$ such that $\lambda_{\min}(\sum_{i \in S_t} x_i x_i^\top) > 1 - 3\varepsilon$ is satisfied so we are done so that Problem (2.2) is solved, or we can always find pairs $(i_t, j_t)$ satisfying Eq. (2.10), which together with Eq. (2.7) implies

$$-\lambda_{\min}\left(\sum_{j \in S_T} x_j x_j^\top\right) \leq \sum_{t=0}^{T-1} -\frac{\varepsilon}{k} + \frac{2\sqrt{p}}{\alpha} = -\frac{T\varepsilon}{k} + 2\varepsilon .$$

Here in the last inequality we apply the choice $\alpha = \sqrt{p}/\varepsilon$. In other words, as long as $T \geq k/\varepsilon$, it must satisfy $\lambda_{\min}(\sum_{i \in S_T} x_i x_i^\top) \geq 1 - 2\varepsilon$, and hence Problem (2.2) is also solved.

**Pseudocode.** We summarize the aforementioned swapping algorithm in Algorithm 1. We also present a binary search routine FINDCONSTANT in Algorithm 2 to compute the constant $c_t \in \mathbb{R}$ so that $\mathrm{tr}(A_t)$ is sufficiently close to 1.[3] In theory, one can show that it suffices to compute $c_t$ up to an additive error $\delta = \Theta(1/\mathrm{poly}(\varepsilon^{-1}, T)) = \Theta(1/\mathrm{poly}(\varepsilon^{-1}, k))$, and our regret inequality (2.7) is not affected by more than an additive $\varepsilon$ on both sides.[4] In practice, we find it sufficient to run binary search $c_t$ for 20 iterations when $p = 50$ and $n = 1000$.

---

[3]Indeed, the maximum possible value for $c_t$ is $c_u = \sqrt{p}$ because $\mathrm{tr}[(\sqrt{p}I + \alpha Z)^{-2}] \geq \mathrm{tr}[(\sqrt{p}I)^{-2}] = 1$, and the minimum possible value of $c_t$ must be above $c_\ell = -\alpha \lambda_{\min}(Z)$, because $\mathrm{tr}[(c_\ell I + \alpha Z)^{-2}] \geq \mathrm{tr}[(\sqrt{p}I)^{-2}] = +\infty$. One can show (see the proof of Claim 2.4) that the value $\mathrm{tr}[(cI + \alpha Z)^{-2}]$ is a monotone function in $c \in (c_\ell, c_u]$.

[4]A careful stability analysis for the $\ell_{1/2}$ strategy has already appeared in the appendix of [4].

**Algorithm 1** A swapping algorithm for rounding

**Input:** design pool $X \in \mathbb{R}^{n \times p}$, a fraction solution $\pi \in [0,1]^n$ with $\|\pi\|_1 = k \geq 5p/\varepsilon^2$, and desired accuracy $\varepsilon > 0$.
**Output:** $\widehat{s} \in \{0,1\}^n$ with $\|\widehat{s}\|_1 = k$ and $\sum_i \widehat{s}_i \cdot x_i x_i^\top \succeq (1 - 3\varepsilon) \sum_i \pi_i \cdot x_i x_i^\top$.
1: $\alpha \leftarrow \sqrt{p}/\varepsilon$ and $T \leftarrow k/\varepsilon$;
2: $X \leftarrow X(X^\top \mathrm{diag}(\pi)X)^{-1/2}$; ▷ whitening step
3: $S_0 \subseteq [n]$ an arbitrary subset of size $k$ and $t \leftarrow 1$; ▷ initialization
4: **while** $t \leq T$ and $\lambda_{\min}(\sum_{i \in S_{t-1}} x_i x_i^\top) \leq 1 - 3\varepsilon$ **do**
5: $\quad c_t \leftarrow \textsc{FindConstant}(\sum_{i \in S_{t-1}} x_i x_i^\top, \alpha)$;
6: $\quad A_t \leftarrow (c_t I + \alpha \sum_{i \in S_{t-1}} x_i x_i^\top)^{-2}$;
7: $\quad i_t \leftarrow \arg\min_{i \in S_{t-1}, 2\alpha \langle A_t^{1/2}, x_i x_i^\top \rangle < 1} \left\{ \frac{\langle A_t, x_i x_i^\top \rangle}{1 - 2\alpha \langle A_t^{1/2}, x_i x_i^\top \rangle} \right\}$;
8: $\quad j_t \leftarrow \arg\max_{j \in [n] \setminus S_{t-1}} \left\{ \frac{\langle A_t, x_j x_j^\top \rangle}{1 + 2\alpha \langle A_t^{1/2}, x_j x_j^\top \rangle} \right\}$;
9: $\quad S_t = S_{t-1} \cup \{j_t\} \setminus \{i_t\}$ and $t \leftarrow t+1$;
10: **end while**
11: **return** $\widehat{s} \in \{0,1\}^n$ where $\widehat{s}_i = 1$ iff $i \in S_{t-1}$.

**Algorithm 2** $\textsc{FindConstant}(Z, \alpha)$

1: $c_\ell = -\alpha \lambda_{\min}(Z)$, $c_u = \sqrt{p}$; $\delta = \Theta(1/\mathsf{poly}(\varepsilon^{-1}, k))$; ▷ initialization
2: **while** $|c_\ell - c_u| > \delta$ **do**
3: $\quad \bar{c} \leftarrow (c_\ell + c_u)/2$;
4: $\quad$ If $\mathrm{tr}[(\bar{c}I + \alpha Z)^{-2}] > 1$ then $c_\ell \leftarrow \bar{c}$; else $c_u \leftarrow \bar{c}$;
5: **end while**
6: **return** $c = (c_\ell + c_u)/2$.

**Time complexity.** We first argue that each run of $\textsc{FindConstant}(Z, \alpha)$ costs $\widetilde{O}(p^3)$ arithmetic operations, where in the $\widetilde{O}(\cdot)$ notation we logarithmic factors in $p$ and $\varepsilon^{-1}$. Indeed, it takes at most $O(p^3)$ arithmetic operations to invert a $p$-dimensioan matrix or compute its eigendecomposition. As for the binary search on $c_t$, it terminates in $O(\log((c_u - c_\ell)/\delta)) = O(\log((\alpha + \sqrt{p})/\delta)) = \widetilde{O}(1)$ iterations.

In the preparation step of Algorithm 1, we can for instance compute the matrix square root $X(X^\top \mathrm{diag}(\pi)X)^{-1/2}$ explicitly in $O(p^3)$ time. In each iteration of Algorithm 1, once $A_t$ and $A_t^{1/2}$ are computed, it takes $O(np^2)$ additional time to search for the candidate indices $i_t$ and $j_t$.

Because $n \geq p$, the overall time complexity of Algorithm 1 is $\widetilde{O}(nkp^2/\varepsilon)$, which is linear in the pool size $n$ when both $k$ and $p$ are small. Such time complexity is not optimal, but further improvements in running time is out of the consideration of this paper. (In particular, we are aware of a warm-restarting variant of Algorithm 1 which achieves time complexity $\widetilde{O}(nkp^2)$.)

Finally, although Algorithm 1 is described for $b = 1$, as we discussed at the beginning of this section, if $b > 1$ we can simply duplicate each $x_i$ exactly $b$ times and then invoke Algorithm 1. We emphasize that the running time of does not blow up by a factor of $b$, because in each iteration, it suffices to go through at most $n$ choices of indices $j_t$ with distinct $x_j$, in order to find the best $j_t$.

## 2.4 Main Regret Lemma: Proof of Lemma 2.5

To prove this lemma we consider an equivalent "2-step" description of the mirror descent procedure:

$$\widetilde{A}_t = \arg\min_{A \succeq 0} \{\Delta_\psi(A_{t-1}, A) + \alpha\langle F_{t-1}, A\rangle\}; \qquad A_t = \arg\min_{A \in \Delta_{p\times p}} \Delta_\psi(\widetilde{A}_t, A).$$

By the so-called "tweaked analysis" of mirror descent [39, 47], the matrix $A_t$ defined above is identical to its original definition of $\arg\min_{A \in \Delta_{p\times p}}\{\Delta_\psi(A_{t-1}, A) + \alpha\langle F_{t-1}, A\rangle\}$. This can also be verified by writing $\widetilde{A}_t$ explicitly using the following claim, and verifying that $A_t$ (in its closed form by Claim 2.4) is indeed a minimizer of $\Delta_\psi(\widetilde{A}_t, A)$ over $A \in \Delta_{p\times p}$ by taking its gradient.

**Claim 2.9.** *We have* $\widetilde{A}_t = (A_{t-1}^{-1/2} + \alpha F_{t-1})^{-2}$.

*Proof.* We first show $(A_{t-1}^{-1/2} + \alpha F_{t-1})^{-2}$ is well defined. By assumption $\alpha\langle A_{t-1}^{1/2}, v_{t-1}v_{t-1}^\top\rangle < 1$, and hence $-\alpha F_{t-1} \preceq \alpha v_{t-1}v_{t-1}^\top \prec A_{t-1}^{-1/2}$. This is because for any matrices $A \succ 0$ and $B \succeq 0$, we have $\langle A, B\rangle = \text{tr}(A^\top B) < 1 \implies A^\top B \preceq I \implies B \preceq A^{-1}$. Consequently, we have $A_{t-1}^{-1/2} + \alpha F_{t-1} \succ 0$ and its inverse exists.

Next, to prove $\widetilde{A}_t = (A_{t-1}^{-1/2} + \alpha F_{t-1})^{-2}$ is a minimizer of the convex function $\{\Delta_\psi(A_{t-1}, A) + \alpha\langle F_{t-1}, A\rangle\}$ over all positive semi-definite matrices $A$, we show its gradient evaluated at $\widetilde{A}_t$ is zero.[5] Indeed,

$$\nabla\left(\Delta_\psi(A_{t-1}, \widetilde{A}_t) + \alpha\langle F_{t-1}, \widetilde{A}_t\rangle\right) = \nabla\psi(\widetilde{A}_t) - \nabla\psi(A_{t-1}) + \alpha F_{t-1}$$

$$= -\widetilde{A}_t^{-1/2} + A_{t-1}^{-1/2} + \alpha F_{t-1} = 0.\qquad\square$$

Since $\nabla\psi(\widetilde{A}_t) - \nabla\psi(A_{t-1}) + \alpha F_{t-1} = 0$ as shown in the proof above, we have (by defining $\widetilde{A}_0 = A_0$)

$$\langle \alpha F_{t-1}, A_{t-1} - U\rangle = \langle \nabla\psi(A_{t-1}) - \nabla\psi(\widetilde{A}_t), A_{t-1} - U\rangle$$
$$= \Delta_\psi(A_{t-1}, U) - \Delta_\psi(\widetilde{A}_t, U) + \Delta_\psi(\widetilde{A}_t, A_{t-1})$$
$$\leq \Delta_\psi(\widetilde{A}_{t-1}, U) - \Delta_\psi(\widetilde{A}_t, U) + \Delta_\psi(\widetilde{A}_t, A_{t-1}). \qquad (2.11)$$

Above, the second equality and the last inequality follow from the "three-point" equality and the generalized Pythagorean theorem of Bregman divergence (see for example, Lemma 2.1 of [4]). Expanding $\Delta_\psi(\widetilde{A}_t, A_{t-1})$ by its definition gives

$$\Delta_\psi(\widetilde{A}_t, A_{t-1}) = \psi(A_{t-1}) - \psi(\widetilde{A}_t) - \langle\nabla\psi(\widetilde{A}_t), A_{t-1} - \widetilde{A}_t\rangle$$
$$= -2\text{tr}(A_{t-1}^{1/2}) + 2\text{tr}(\widetilde{A}_t^{1/2}) + \langle\widetilde{A}_t^{-1/2}, A_{t-1} - \widetilde{A}_t\rangle$$
$$= \langle\widetilde{A}_t^{-1/2}, A_{t-1}\rangle + \text{tr}(\widetilde{A}_t^{1/2}) - 2\text{tr}(A_{t-1}^{1/2})$$
$$= \langle A_{t-1}^{-1/2} + \alpha F_{t-1}, A_{t-1}\rangle + \text{tr}(\widetilde{A}_t^{1/2}) - 2\text{tr}(A_{t-1}^{1/2})$$
$$= \alpha\langle F_{t-1}, A_{t-1}\rangle + \text{tr}(\widetilde{A}_t^{1/2}) - \text{tr}(A_{t-1}^{1/2}). \qquad (2.12)$$

---
[5]The convexity of this objective follows from Lieb's concavity theorem [12, 33], and is already a known fact in matrix regret minimization literatures [4].

Combining Eqs. (2.11) and (2.12) and telescoping from $t = 1$ to $t = T$ we obtain

$$-\alpha \sum_{t=0}^{T-1} \langle F_t, U \rangle \leq \Delta_\psi(A_0, U) - \Delta_\psi(\widetilde{A}_T, U) + \sum_{t=0}^{T-1} \text{tr}(\widetilde{A}_{t+1}^{1/2}) - \text{tr}(A_t^{1/2})$$

$$\leq \Delta_\psi(A_0, U) + \sum_{t=0}^{T-1} \text{tr}(\widetilde{A}_{t+1}^{1/2}) - \text{tr}(A_t^{1/2}), \quad (2.13)$$

where the second inequality holds because Bregman divergence $\Delta_\psi(\widetilde{A}_T, U)$ is always non-negative.

It remains to upper bound the "consecutive difference" $\text{tr}(\widetilde{A}_{t+1}^{1/2}) - \text{tr}(A_t^{1/2})$.

Let $P_t = \sqrt{\alpha}[u_t \ v_t] \in \mathbb{R}^{p \times 2}$ and $J = \text{diag}(1, -1) \in \mathbb{R}^{2 \times 2}$, so we have $\alpha F_t = P_t J P_t^\top$. By the definition of $\widetilde{A}_{t+1}^{1/2}$ and the Woodbury formula[6],

$$\text{tr}(\widetilde{A}_{t+1}^{1/2}) = \text{tr}\left[(A_t^{-1/2} + P_t J P_t^\top)^{-1}\right] = \text{tr}\left[A_t^{1/2} - A_t^{1/2} P_t (J + P_t^\top A_t^{1/2} P_t)^{-1} P_t^\top A_t^{1/2}\right]. \quad (2.14)$$

It is crucial to spectrally lower bound the core $2 \times 2$ matrix $(J + P_t^\top A_t^{1/2} P_t)^{-1/2}$ in the middle of Eq. (2.14). For this purpose, we claim that

**Claim 2.10.** *Suppose $P_t^\top A_t^{1/2} P_t = [b \ d; d \ c] \in \mathbb{R}^{2 \times 2}$ and $2\alpha \langle A_t^{1/2}, v_t v_t^\top \rangle < 1$. Then*

$$(J + P_t^\top A_t^{1/2} P_t)^{-1} = \left(J + \begin{bmatrix} b & d \\ d & c \end{bmatrix}\right)^{-1} \succeq \left(J + \begin{bmatrix} 2b & 0 \\ 0 & 2c \end{bmatrix}\right)^{-1}.$$

Claim 2.10 is trivially true if $J \succeq 0$, but becomes less obvious when $J$ has negative eigenvalues. In fact, Claim 2.10 is not universally true for any matrices of the form $PAP^\top$, and specifically requires the condition that $2\alpha \langle A_t^{1/2}, v_t v_t^\top \rangle < 1$. We defer the proof of Claim 2.10 to the appendix.

With Claim 2.10, the consecutive gap $\text{tr}(\widetilde{A}_{t+1}^{1/2}) - \text{tr}(A_t^{1/2})$ can be bounded as

$$\text{tr}(\widetilde{A}_{t+1}^{1/2}) - \text{tr}(A_t^{1/2}) = -\text{tr}\left[-A_t^{1/2} P_t (J + P_t^\top A_t^{1/2} P_t)^{-1} P_t^\top A_t^{1/2}\right]$$

$$\leq -\text{tr}\left[-A_t^{1/2} P_t \left(J + \begin{bmatrix} 2\alpha u_t^\top A_t^{1/2} u_t & 0 \\ 0 & 2\alpha v_t^\top A_t^{1/2} v_t \end{bmatrix}\right)^{-1} P_t^\top A_t^{1/2}\right]$$

$$= -\frac{\alpha \langle A_t, u_t u_t^\top \rangle}{1 + 2\alpha \langle A_t^{1/2}, u_t u_t^\top \rangle} + \frac{\alpha \langle A_t, v_t v_t^\top \rangle}{1 - 2\alpha \langle A_t^{1/2}, v_t v_t^\top \rangle}. \quad (2.15)$$

Plugging Eq. (2.15) into Eq. (2.13) we complete the proof of Lemma 2.5.

## 2.5 Main Averaging Lemma: Proof of Lemma 2.8

**Lemma 2.8** (main averaging lemma). *For every $\varepsilon > 0$ and subset $S \subseteq [n]$ of cardinality $k$, suppose $\lambda_{\min}(\sum_{i \in S} x_i x_i^\top) \leq 1 - 3\varepsilon$ and $A = (cI + \alpha \sum_{i \in S} x_i x_i^\top)^{-2}$, where $c \in \mathbb{R}$ is the unique number such that $A \succeq 0$ and $\text{tr}(A) = 1$. Then, the following statements are true:*

$$\nu := \min_{i \in S, 2\alpha \langle A^{1/2}, x_i x_i^\top \rangle < 1} \frac{\langle A, x_i x_i^\top \rangle}{1 - 2\alpha \langle A^{1/2}, x_i x_i^\top \rangle} \leq \frac{1 - \varepsilon}{k}; \quad (2.8)$$

---

[6]$(A + UCV)^{-1} = A^{-1} - A^{-1}U(C^{-1} + VA^{-1}U)^{-1}VA^{-1}$, provided that all inverses exist.

$$\max_{j \in [n] \setminus S} \frac{\langle A, x_j x_j^\top \rangle}{1 + 2\alpha \langle A^{1/2}, x_j x_j^\top \rangle} \geq \nu + \frac{\varepsilon}{k}. \tag{2.9}$$

*Furthermore, if $\alpha = \sqrt{p}/\varepsilon$ and $k \geq 5p/\varepsilon^2$ for some $\varepsilon \in (0, 1/3]$, then there always exists $i \in S$ such that $2\alpha \langle A^{1/2}, x_i x_i^\top \rangle < 1$.*

We first state a technical claim as follows:

**Claim 2.11.** *Suppose $Z \succeq 0$ is a p-dimensional PSD matrix with $\lambda_{\min}(Z) \leq 1$. Let $A = (\alpha Z + cI)^{-2}$, where $c \in \mathbb{R}$ is the unique real number such that $A \succeq 0$ and $\mathrm{tr}(A) = 1$. Then*

1. $\alpha \langle A^{1/2}, Z \rangle \leq p + \alpha \sqrt{p}$;
2. $\langle A, Z \rangle \leq \sqrt{p}/\alpha + \lambda_{\min}(Z)$.

The proof of Claim 2.11 is by writing $A$ and $Z$ as diagonal matrices in their common eigen basis, and then applying Cauchy-Schwarz. We defer it to the appendix.

Back to the proof of Lemma 2.8, we first show the existence of (at least one) $i \in S$ such that $2\alpha \langle A^{1/2}, x_i x_i^\top \rangle < 1$. Define $Z = \sum_{i \in S} x_i x_i^\top$, and by definition $A = (cI + \alpha \sum_{i \in S} x_i x_i^\top)^{-2} = (\alpha Z + cI)^{-2}$. Assume by way of contradiction that such $i$ does not exist. We then have

$$\sum_{i \in S} 2\alpha \langle A^{1/2}, x_i x_i^\top \rangle = 2\alpha \langle A^{1/2}, Z \rangle \geq |S| = k. \tag{2.16}$$

On the other hand, because $Z \succeq 0$ and $\lambda_{\min}(Z) < 1$, invoking Claim 2.11 we get

$$2\alpha \langle A^{1/2}, Z \rangle \leq 2p + 2\alpha \sqrt{p}$$

which contradicts Eq. (2.16) provided that $\alpha = \sqrt{p}/\varepsilon$ and $k > 4p/\varepsilon$. Thus, there must exist $i \in S$ such that $2\alpha \langle A^{1/2}, x_i x_i^\top \rangle < 1$.

In fact, the same "proof by contradiction" also implies $\sum_{i \in S}(1 - 2\alpha \langle A^{1/2}, x_i x_i^\top \rangle) > 0$.

**Proof of Eq. (2.8).** By definition of $\nu$, we must have that

$$(1 - 2\alpha \langle A^{1/2}, x_i x_i^\top \rangle)\nu \leq \langle A, x_i x_i^\top \rangle \quad \text{for all} \quad i \in S \ ,$$

because if $2\alpha \langle A^{1/2}, x_i x_i^\top \rangle \geq 1$ the left-hand side is non-positive while the right-hand side is always non-negative, thanks to the positive semi-definiteness of $A$. Subsequently,

$$\nu \overset{①}{\leq} \frac{\sum_{i \in S} \langle A, x_i x_i^\top \rangle}{\sum_{i \in S}(1 - 2\alpha \langle A^{1/2}, x_i x_i^\top \rangle)} \overset{②}{\leq} \frac{\sqrt{p}/\alpha + \lambda_{\min}(\sum_{i \in S} x_i x_i^\top)}{k - 2p - 2\alpha \sqrt{p}} \overset{③}{\leq} \frac{\varepsilon + 1 - 3\varepsilon}{k(1 - 8\varepsilon/15)} \overset{④}{\leq} \frac{1 - \varepsilon}{k} \ .$$

Above, inequality ① holds because the denominator is strictly positive as we have shown; inequality ② is due to Claim 2.11; inequality ③ has used our choices $\alpha = \sqrt{p}/\varepsilon, k \geq 5p/\varepsilon^2, \varepsilon \leq 1/3$, and our assumption $\lambda_{\min}(\sum_{i \in S} x_i x_i^\top) \leq 1 - 3\varepsilon$; and inequality ④ has used $\varepsilon \leq 1/3$. We have thus proved that $\nu \leq (1 - \varepsilon)/k$.

**Proof of Eq. (2.9).** Define $t = \nu + \varepsilon/k \leq 1/k$. To prove Eq. (2.9) it suffices to show that

$$\sum_{j \in [n] \setminus S} \pi_j \langle A, x_j x_j^\top \rangle \geq t \sum_{j \in [n] \setminus S} \pi_j (1 + 2\alpha \langle A^{1/2}, x_j x_j^\top \rangle) \ , \tag{2.17}$$

because $\pi_j \geq 0$ for all $j$. Recall that $\sum_{j=1}^n \pi_j = k, \sum_{j=1}^n \pi_j x_j x_j^\top = I$. We then have

$$\sum_{j \in [n] \setminus S} \pi_j (1 + 2\alpha \langle A^{1/2}, x_j x_j^\top \rangle) = \left(k - \sum_{j \in S} \pi_j\right) + 2\alpha \cdot \sum_{j \in [n] \setminus S} \pi_j \langle A^{1/2}, x_j x_j^\top \rangle$$

$$\leq \left(k - \sum_{j \in S} \pi_j\right) + 2\alpha \cdot \sum_{j=1}^{n} \pi_j \langle A^{1/2}, x_j x_j^\top \rangle$$

$$= k - \sum_{j \in S} \pi_j + 2\alpha \langle I, A^{1/2} \rangle = k - \sum_{j \in S} \pi_j + 2\alpha \operatorname{tr}(A^{1/2}).$$

Similarly,

$$\sum_{j \in [n] \setminus S} \pi_j \langle A, x_j x_j^\top \rangle = \left\langle I - \sum_{j \in S} \pi_j x_i x_i^\top, X \right\rangle = \operatorname{tr}(A) - \sum_{j \in S} \pi_j \langle A, x_j x_j^\top \rangle.$$

Note that for any $p \times p$ positive semi-definite matrix $Z \succeq 0$, $\operatorname{tr}(Z^{1/2}) \leq \sqrt{p \cdot \operatorname{tr}(Z)}$ thanks to the Hölder's inequality[7] applied to the non-negative spectrum of $Z^{1/2}$, and that $\operatorname{tr}(A) = 1$ by definition. Subsequently,

$$\sum_{j \in [n] \setminus S} \pi_j \langle A, x_j x_j^\top \rangle - t \cdot \sum_{j \in [n] \setminus S} \pi_j (1 + 2\alpha \langle A^{1/2}, x_j x_j^\top \rangle)$$

$$\geq \operatorname{tr}(A) - \sum_{j \in S} \pi_j \langle A, x_j x_j^\top \rangle - t \left(k - \sum_{j \in S} \pi_j\right) - 2\alpha t \cdot \operatorname{tr}(A^{1/2})$$

$$\geq 1 - \sum_{j \in S} \pi_j \langle A, x_j x_j^\top \rangle - t \left(k - \sum_{j \in S} \pi_j\right) - 2\alpha t \sqrt{p}$$

$$= 1 - tk - 2t\alpha\sqrt{p} - \sum_{j \in S} \pi_j (\langle A, x_j x_j^\top \rangle - t)$$

$$\overset{①}{\geq} 1 - tk - 2t\alpha\sqrt{p} - \sum_{j \in S} \max\{\langle A, x_j x_j^\top \rangle - t, 0\}$$

$$\geq 1 - tk - 2t\alpha\sqrt{p} - \sum_{j \in S} (\langle A, x_j x_j^\top \rangle - t) - \sum_{j \in S} \max\{(t - \langle A, x_j x_j^\top \rangle), 0\}$$

$$\overset{②}{\geq} 1 - 2t\alpha\sqrt{p} - \sqrt{p}/\alpha - \lambda_{\min}\left(\sum_{j \in S} x_j x_j^\top\right) - \sum_{j \in S} \max\{(t - \langle A, x_j x_j^\top \rangle), 0\} \ . \quad (2.18)$$

Above, inequality ① holds because $\pi_j \leq 1$ for all $j$; and in inequality ② we have applied $\sum_{j \in S} \langle A, x_j x_j^\top \rangle \leq \sqrt{p}/\alpha + \lambda_{\min}(\sum_{j \in S} x_j x_j^\top)$ which comes from Claim 2.11.

By the conditions that $\alpha = \sqrt{p}/\varepsilon$, $t \leq 1/k$ and $\lambda_{\min}(\sum_{j \in S} x_j x_j^\top) \leq 1 - 3\varepsilon$, we can simplify the right-hand side of Eq. (2.18) and thus obtain

$$\sum_{j \in [n] \setminus S} \pi_j \left\{ \langle A, x_j x_j^\top \rangle - t(1 + 2\alpha \langle A^{1/2}, x_j x_j^\top \rangle) \right\} \geq 2\varepsilon - \frac{2p}{\varepsilon k} - \sum_{j \in S} \max\{t - \langle A, x_j x_j^\top \rangle, 0\}.$$
(2.19)

Furthermore, because $(1 - 2\alpha \langle A^{1/2}, x_i x_i^\top \rangle)\nu \leq \langle A, x_i x_i^\top \rangle$ for all $i \in S$, invoking Claim 2.11 we

---

[7]$|x_1| + \cdots + |x_d| \leq \sqrt{d} \cdot \sqrt{x_1^2 + \cdots + x_d^2}$ for any sequences of $d$ real numbers $x_1, \ldots, x_n$.

have
$$\sum_{i \in S'} (\nu - \langle A, x_i x_i^\top \rangle) \leq \sum_{i \in S'} 2\nu\alpha \langle A^{1/2}, x_j x_j^\top \rangle \leq 2\nu(p + \alpha\sqrt{p}) \quad \text{for all } S' \subseteq S.$$

Consider $S' = \{i \in S : t - \langle A, x_i x_i^\top \rangle \geq 0\}$. We then have
$$\sum_{j \in S'} \max\{t - \langle A, x_j x_j^\top \rangle, 0\} = \sum_{j \in S'} (t - \langle A, x_j x_j^\top \rangle) = (t-\nu)|S'| + \sum_{j \in S'} (\nu - \langle A, x_j x_j^\top \rangle)$$
$$\leq (t-\nu)k + 2\nu(p + \alpha\sqrt{p}) \leq \varepsilon + \frac{3p/\varepsilon}{k}, \quad (2.20)$$

where the last two inequalities hold because $t - \nu = \varepsilon/k \geq 0$, $|S'| \leq |S| = k$, $\nu \leq 1/k$ and $\alpha = \sqrt{p}/\varepsilon$. Combining Eqs. (2.19) and (2.20) we arrive at

$$\sum_{j \in [n] \setminus S} \pi_j \left\{ \langle A, x_j x_j^\top \rangle - t(1 + 2\alpha \langle A^{1/2}, x_j x_j^\top \rangle) \right\} \geq \varepsilon - \frac{5p}{\varepsilon k}.$$

If $k \geq 5p/\varepsilon^2$, the right-hand side of the above inequality is non-negative, which finishes the proof of Eq. (2.17) and thus also the proof of Eq. (2.9).

## 3 Solving the Continuous Relaxation

In this section we address the problem of efficiently solving the continuous relaxed problem in Eq. (1.3) which we repeat here:

$$\min_{s \in \mathcal{C}_{k,b}} F(s) = \min_{s \in \mathcal{C}_{k,b}} f\left(\sum_{i=1}^n s_i \cdot x_i x_i^\top\right) \quad \text{where} \quad \mathcal{C}_{k,b} := \{s \in [0,b]^d : \sum_{i=1}^n s_i \leq k\}.$$

up to arbitrary precision. While our primary considerations are A/D/T/E/V/G optimality, we state our results in the most general form so as to be applicable to other optimality criteria.

The optimization algorithm we analyze in this section is the *entropic mirror descent* method (see for instance [11]). While other methods such as projected gradient descent or conic programming [16] are also applicable, we recommend entropic mirror descent because it suits the geometry of our problem and is also observed to converge fast in simulations.

### 3.1 Properties and Assumptions

Let $\pi^* \in \mathrm{argmin}_{s \in \mathcal{C}_{k,b}} F(x)$ be an optimal fractional solution to Eq. (1.3). Since $\mathcal{C}_{k,b} \subseteq \mathcal{S}_{k,b}$, we know that any integral solution $s \in \mathcal{S}_{k,b}$ is also feasible to Eq. (1.3), and therefore

**Proposition 3.1.** $F(\pi) \leq \min_{s \in \mathcal{S}_{k,b}} F(s)$.

We also have the following simple property:

**Proposition 3.2.** *There exists an optimal solution $\pi^*$ to Eq. (1.3) with $\sum_{i=1}^n \pi_i^* = k$.*

*Proof.* Suppose $\sum_{i=1}^n \pi_i^* < k$. Then since $nb \geq k$, there must exist $i \in [n]$ such that $\pi_i^* < b$. Define $\widetilde{\pi}_j^* = \pi_j^*$ for $j \neq i$ and $\widetilde{\pi}_i^* = \min\{b, \pi_i^* + k - \|\pi^*\|_1\} > \pi_i^*$. It is clear that $\widetilde{\pi}^*$ is feasible and $(\sum_{i=1}^n \pi_i^* x_i x_i^\top) \preceq (\sum_{i=1}^n \widetilde{\pi}_i^* x_i x_i^\top)$. By monotonicity of $f$, we have $f(\sum_{i=1}^n \pi_i^* x_i x_i^\top) \geq f(\sum_{i=1}^n \widetilde{\pi}_i^* x_i x_i^\top)$ and hence $\widetilde{\pi}^*$ is also optimal. Repeat this procedure until $\sum_{i=1}^n \widetilde{\pi}_i^* = k$. □

To efficiently solve the continuous relaxation, we impose the following assumptions on $f$:

(B1) $f$ is convex: $f(tA + (1-t)B) \leq tf(A) + (1-t)f(B)$ for all $A, B \in \mathbb{S}_p^+$ and $t \in [0,1]$;

(B2) For any $\lambda > 0$, there exists parameter $L_\lambda > 0$ such that the "smoothed" objective
$$F_\lambda \colon \pi \mapsto f(\textstyle\sum_{i=1}^n (\pi_i + \lambda/n) x_i x_i^\top)$$
is $L_\lambda$-Lipschitz continuous in $\|\cdot\|_1$: that is, $|F_\lambda(\pi) - F_\lambda(\pi')| \leq L_\lambda \|\pi - \pi'\|_1$ for all $\pi, \pi' \in \mathcal{C}_{b,k}$;

(B3) There exists $\mu_0 > 0$ such that $\inf_{\pi \in \mathcal{C}_{b,k}} F(\pi) \geq \mu_0$.

*Remark* 3.3. Suppose $\underline{\sigma} I_p \preceq \frac{1}{n} \sum_{i=1}^n x_i x_i^\top \preceq \overline{\sigma} I_p$ for some $0 < \underline{\sigma} \leq \overline{\sigma} < \infty$ and $\max_{1 \leq i \leq n} \|x_i\|_2 \leq B$. Then all A/T/E/V/G optimality satisfy (B1) through (B3) with the following parameters:

| | | | |
|---|---|---|---|
| $f_A : \Sigma \mapsto \mathrm{tr}(\Sigma^{-1})/p$ | satisfies | $L_\lambda = B^2/\lambda^2 \underline{\sigma}^2 p,$ | $\mu_0 = 1/kb\overline{\sigma};$ |
| $f_T : \Sigma \mapsto p/\mathrm{tr}(\Sigma)$ | satisfies | $L_\lambda = B^2/\lambda \underline{\sigma},$ | $\mu_0 = 1/kb\overline{\sigma};$ |
| $f_E : \Sigma \mapsto \|\Sigma^{-1}\|_2$ | satisfies | $L_\lambda = B^2/\lambda^2 \underline{\sigma}^2,$ | $\mu_0 = 1/kb\overline{\sigma};$ |
| $f_V : \Sigma \mapsto \mathrm{tr}(X \Sigma^{-1} X^\top)$ | satisfies | $L_\lambda = B^2 \overline{\sigma}/\lambda^2 \underline{\sigma}^2,$ | $\mu_0 = \underline{\sigma}/kb\overline{\sigma};$ |
| $f_G : \Sigma \mapsto \max \mathrm{diag}(X \Sigma^{-1} X^\top)$ | satisfies | $L_\lambda = B^2/\lambda^2 \underline{\sigma}^2,$ | $\mu_0 = B^2/kb\overline{\sigma}.$ |

One notable exception is the D-optimality $f_D \colon \Sigma \mapsto (\det \Sigma)^{-1/p}$, which does not satisfy (B1) because $f_D$ is not convex. For this particular objective, it is a well-established practice to consider the negative *log-determinant* $\log f_D \colon \Sigma \mapsto -\frac{1}{p} \log \det \Sigma$, which is convex (see for example [9]). This motivates us to consider an alternative set of assumptions that concern the $\log f$ function:

(C1) $\log f$ is convex, meaning that $\log f(tA + (1-t)B) \leq t \log f(A) + (1-t) \log f(B)$ for all $A, B \in \mathbb{S}_p^+$ and $t \in [0,1]$;

(C2) For any $\lambda > 0$, there exists a parameter $L_\lambda > 0$ such that
$$\log F_\lambda \colon \pi \mapsto \log f(\textstyle\sum_{i=1}^n (\pi_i + \lambda/n) x_i x_i^\top)$$
is $L_\lambda$-Lipschitz continuous with respecrt to $\|\cdot\|_1$, meaning that $|\log F_\lambda(\pi) - \log F_\lambda(\pi')| \leq L_\lambda \|\pi - \pi'\|_1$ for all $\pi, \pi' \in \mathcal{C}_{b,k}$.

*Remark* 3.4. Suppose $\underline{\sigma} I_p \preceq \frac{1}{n} \sum_{i=1}^n x_i x_i^\top \preceq \overline{\sigma} I_p$ for some $0 < \underline{\sigma} \leq \overline{\sigma} < \infty$ and $\max_{1 \leq i \leq n} \|x_i\|_2 \leq B$. The D-optimality criterion

$f_D : \Sigma \mapsto (\det \Sigma)^{-1/p}$   satisfies (C1) and (C2) with   $L_\lambda = B^2/\lambda \underline{\sigma} p,$   $\mu_0 = 1/kb\overline{\sigma}.$

*Remark* 3.5. The Bayesian experimental design criteria satisfy (B1) through (B3) as well (except for the D-optimality). Since they add a multiple of the identity matrix to the covariance matrix (see Section 1.1), they satisfy (B2) with parameter $L_\lambda$ *independent* of $\lambda$, meaning that the Lipschitz continuity holds for all $\lambda > 0$. Similarly, the Bayesian version of $f_D$ also satisfies (C1) through (C2).

**Algorithm 3** The projected entropic mirror descent algorithm for solving Eq. (3.2).

**Input:** function $\min_\omega \widetilde{F}_\lambda(\omega)$ defined in Eq (3.2); its Lipschitz constant $\widetilde{L}_\lambda$; and $T$ number of iterations.
1: $\omega^{(0)} = (1/n, \cdots, 1/n)$; ▷ initialization
2: **for** $t \leftarrow 0$ **to** $T-1$ **do**
3:     $\eta_t \leftarrow \Theta(\widetilde{L}_\lambda^{-1}\sqrt{\log n/(t+1)})$ ▷ learning rate
4:     Compute subgradient $g^{(t)} \in \partial \widetilde{F}_\lambda(\omega^{(t)})$;
5:     Update: $\omega_i^{(t+1/2)} \propto \omega_i^{(t)}\exp\{-\eta_t g_i^{(t)}\}$, normalized so that $\omega^{(t+1/2)} \in \Delta_n$;
6:     Projection: $\omega^{(t+1)} \leftarrow \text{BoxSimplexProject}(\omega^{(t+1/2)}, b/k)$; ▷ see Algorithm 4
7: **end for**
8: **return** $\omega_{\lambda,t} := \frac{1}{T}\sum_{t=0}^{T-1}\omega^{(t)}$.

## 3.2 Entropic Mirror Descent and its Convergence

Applying Proposition 3.2 and the change-of-variable $\omega = \pi/k$, Eq. (1.3) can be equivalently reformulated as

$$\min_\omega \widetilde{F}(\omega) := \min_\omega f\left(\sum_{i=1}^n k\omega_i x_i x_i^\top\right) \quad \text{s.t.} \quad \omega \in \frac{1}{k}\mathcal{C}_{k,b} := \{\omega \in \Delta_n \mid \|\omega\|_\infty \leq b/k\}, \quad (3.1)$$

where $\Delta_n = \{w \in \mathbb{R}^n : w_i \geq 0, \sum_{i=1}^n w_i = 1\}$ is the simplex of probability vectors.

To avoid singularity issues when $\pi$ is close to 0, we also consider the following "smoothed" version of the optimization problem:

$$\min_\omega \widetilde{F}_\lambda(\omega) := \min_\omega f\left(\sum_{i=1}^n \frac{k}{1+\lambda}(\omega_i + \lambda/n)x_i x_i^\top\right) \quad \text{s.t.} \quad \omega \in \frac{1}{k}\mathcal{C}_{k,b}. \quad (3.2)$$

Here $\lambda \in (0,1)$ is a smoothing parameter that will be specified later. Denote $\omega^*$ and $\omega_\lambda^*$ as the optimal solutions to Eqs. (3.1) and (3.2) respectively. The following proposition establishes the connection between $\widetilde{F}(\omega^*)$ and $\widetilde{F}_\lambda(\omega_\lambda^*)$.

**Proposition 3.6.** *For any $\lambda \in (0,1)$, it holds that $\widetilde{F}(\omega^*) \leq \widetilde{F}_\lambda(\omega_\lambda^*) \leq (1+\lambda)\widetilde{F}(\omega^*)$.*

*Proof.* Construct $\omega_\lambda^\diamond := (1+\lambda)^{-1}(\omega_\lambda^* + \frac{\lambda}{n}\vec{1})$. It is easy to verify that $\omega_\lambda^\diamond \in \frac{1}{k}\mathcal{C}_{k,b}$ because $k \leq bn$. Furthermore, we have $\widetilde{F}(\omega_\lambda^\diamond) = \widetilde{F}_\lambda(\omega_\lambda^*)$. Therefore $\widetilde{F}(\omega^*) \leq \widetilde{F}(\omega_\lambda^\diamond) = \widetilde{F}_\lambda(\omega_\lambda^*)$ by the optimality of $\omega^*$. On the other hand, $\omega^* \in \frac{1}{k}\mathcal{C}_{k,b}$ and thus $\widetilde{F}_\lambda(\omega_\lambda^*) \leq \widetilde{F}_\lambda(\omega^*)$. By monotonicity and sub-linearity of $f$, we have that $\widetilde{F}_\lambda(\omega^*) \leq f(\sum_{i=1}^n \frac{k}{1+\lambda}\omega_i^* x_i x_i^\top) \leq (1+\lambda)f(\sum_{i=1}^n k\omega_i^* x_i x_i^\top) = (1+\lambda)\widetilde{F}(\omega^*)$. Thus, $\widetilde{F}_\lambda(\omega_\lambda^*) \leq \widetilde{F}_\lambda(\omega^*) \leq (1+\lambda)\widetilde{F}(\omega^*)$. □

The *entropic mirror descent* [11] is a classical algorithm that takes into account the geometry of high-dimensional probabilistic simplex to efficiently solve constrained convex optimization problems. At a high level, entropic mirror descent uses the Kullbeck-Leibler (KL) divergence $\sum_i x_i \log(x_i/y_i)$ as the Bregman divergence, whose proximal operator can be evaluated in closed form as multiplicative weight updates. We describe in Algorithm 3 how (projected) entropic mirror descent is applied to solve the smoothed problem in Eq. (3.1). As our problem has an extra box constraint $\omega_i \leq b/k$, we present in Algorithm 4 a simple algorithm that computes such projection in $O(n \log n)$ time and the KL divergence. The projection algorithm is (in principle similar to but) much simpler than existing algorithms that compute projections onto simplex or $L_1$ balls [25, 29]. The correctness of Algorithm 4 is shown in Appendix A.4.

The following lemma which is an adaptation of Theorem 5.1 in [11] gives the convergence rate of Algorithm 3 when $\widetilde{F}_\lambda$ is Lipschitz continuous:

**Lemma 3.7.** *Fix* $\lambda \in (0, 1)$. *Suppose there exists* $\widetilde{L}_\lambda > 0$ *such that* $|\widetilde{F}_\lambda(a) - \widetilde{F}_\lambda(b)| \leq \widetilde{L}_\lambda \|a - b\|_1$ *holds for all* $a, b \in \frac{1}{k} \mathcal{C}_{k,b}$. *Let* $\omega_\lambda$ *be the output of Algorithm 3 for* $T$ *iterations. Then,*

$$\widetilde{F}_\lambda(\omega_\lambda) - \widetilde{F}_\lambda(\omega_\lambda^*) \leq O\Big(\widetilde{L}_\lambda \sqrt{\frac{\log n}{T}}\Big). \tag{3.3}$$

Suppose the optimality criterion $f$ satisfies (B1) through (B3). It is then easy to verify that $\widetilde{L}_\lambda \leq k L_{\widetilde{\lambda}}$ where $\widetilde{\lambda} = k\lambda$ and $L_{\widetilde{\lambda}}$ is the Lipschitz parameter defined in (B2). Combining Lemma 3.7 and Proposition 3.6 we have the following corollary:

**Corollary 3.8.** *Fix* $\delta \in (0, 1)$. *Suppose (B1) through (B3) hold and we choose* $\lambda = \delta/2$. *Then with* $T = \Omega(\delta^{-2} \mu_0^{-2} L_{k\delta/2}^2 \cdot k^2 \log n)$, *the output of Algorithm 3 satisfies* $\widetilde{F}(\omega_\lambda) \leq (1+\delta)\widetilde{F}(\omega^*)$ *and therefore*

$$k\omega_\lambda \in \mathcal{C}_{k,b} \quad \text{and} \quad F(k\omega_\lambda) \leq (1+\delta)\widetilde{F}(\pi^*) \ .$$

(Recall that the parameters $\mu_0$ and $L_\lambda$ are given in Remark 3.3.)

For the special case of the D-optimality criterion that satisfies (C1) and (C2), an *additive* approximation of $\log F$ implies a relative approximation of $F$. Therefore, we have

**Corollary 3.9.** *Fix arbitrary* $\delta \in (0, 1)$. *Suppose (C1) and (C2) hold, and* $\lambda = \delta/2$. *Then with* $T = \Omega(\delta^{-2} L_{k\lambda/2}^2 \cdot k^2 \log n)$, *the output of Algorithm 3 (on function* $\log \widetilde{F}_\lambda$ *as the objective) satisfies* $\widetilde{F}(\omega_\lambda) \leq (1+\delta)\widetilde{F}(\omega^*)$ *and therefore*

$$k\omega_\lambda \in \mathcal{C}_{k,b} \quad \text{and} \quad F(k\omega_\lambda) \leq (1+\delta)\widetilde{F}(\pi^*) \ .$$

(Recall that the parameter $L_\lambda$ is given in Remark 3.4.)

Corollaries 3.8 and 3.9 show that for optimality criteria that satisfy (B1) through (B3) or (C1) and (C2), an approximate solution to $\pi^*$ with $(1 + \delta)$ relative error can be efficiently computed (1) the subgradients $\nabla \widetilde{F}_\lambda(\omega)$ can be computed efficiently (which is the case for all A/D/T/E/V/G optimality criteria that we study), and (2) parameters $L_\lambda$ and $\mu_0$ are well bounded (see Remark 3.3 and 3.4 and 3.5).

We emphasize that entropic mirror descent does not imply the continuous problem (1.3) is polynomial time solvable, because the parameters $L_\lambda$ and $\mu_0$ depend on the properties of the covariance matrix $\sum_{i=1}^n x_i x_i^\top$ which, in theory, may not be polynomially bounded. One can of course use the ellipsoid method to give a polynomial-time algorithm for (1.3) for theory purpose.[8] However, we

---

[8]The ellipsoid method runs in polynomial time as long as (1) the domain is polynomially bounded (which is the case since the probability simplex is bounded and (2) the separation oracle can be implemented in polynomial time (which is the case for all the optimality criteria that we study in this paper.

**Algorithm 4** Projection onto the probabilistic simplex with box constraint
---
**Input:** $\omega \in \Delta_n$, parameter $b \in [1/n, 1]$.
**Output:** an output $\omega' \in \Delta_n$ such that $\|\omega'\|_\infty \leq b$.      ▷ $\omega' = \arg\min_{y \in \Delta_n, y_i \leq b} \mathrm{KL}(y\|\omega)$
     ▷ where $\mathrm{KL}(y\|\omega) := \sum_i y_i \log \frac{y_i}{\omega_i}$

1: Sort $\omega$ in descending order: $\omega_1 \geq \omega_2 \geq \cdots \geq \omega_n > 0$;
2: **if** $\omega_1 \leq b$ **then return** $\omega$;
3: $\mathrm{KL}_1 \leftarrow b \log(b/\omega_1), \quad Z_1 \leftarrow 1 - \omega_1, \quad \mathrm{KL}_{\mathrm{opt}} \leftarrow \infty, \quad \eta_{\mathrm{opt}} \leftarrow 0, \quad \text{and } C_{\mathrm{opt}} \leftarrow 0$;
4: **for** $q \leftarrow 2$ to $n$ **do**
5:      $C \leftarrow (1 - b(q-1))/Z_{q-1}$;
6:      **if** $C > 0$ and $Cw_q \leq b$ and $\bigl(\mathrm{KL}_{q-1} + C \log(C) \cdot Z_{q-1} \leq \mathrm{KL}_{\mathrm{opt}}\bigr)$ **then**
7:         $\mathrm{KL}_{\mathrm{opt}} \leftarrow \mathrm{KL}_{q-1} + C \log(C) \cdot Z_{q-1}, \quad \eta_{\mathrm{opt}} \leftarrow \omega_q, \quad C_{\mathrm{opt}} \leftarrow C$;
8:      **end if**
9:      $\mathrm{KL}_q \leftarrow b \log(b/\omega_q), \quad Z_q \leftarrow Z_{q-1} - \omega_q$;
10: **end for**
11: Set $\omega'_i \leftarrow b$ if $\omega_i \geq \eta_{\mathrm{opt}}$, and $\omega'_i \leftarrow C_{\mathrm{opt}} \omega_i$ if $\omega_i < \eta_{\mathrm{opt}}$;
12: **return** $\omega'$.

---

still recommend entropic mirror descent because it runs fast in our simulations, see Section 4.[9]

## 4 Empirical Evaluation

We provide some numerical results on Algorithm 1 and compare with popular competitors on the discrete optimization problem for experimental design.

### 4.1 Methods and Our Implementation

The choice $\alpha = \sqrt{p}/\varepsilon$ and the stopping rule in Algorithm 1 are backed by our theoretical proofs, and may be too pessimistic for practical usage. Therefore, we make the following slight changes.

For the choice of $\alpha$, we consider a grid of values $\alpha = \nu\sqrt{p}$ for $\nu = 0.2, 0.4, 0.6, 0.8, 1.0, 1.2, 1.4, 1.6, 1.8, 2.0, 2.5, 3.0, 4.0, 5.0$, and select the output $\widehat{s}$ that leads to the largest $\lambda_{\min}(\sum_i \widehat{s}_i x_i x_i^\top)$. For the stopping rule, we stop the algorithm whenever no $i \in S_{t-1}$ satisfies $2\alpha \langle A_t^{1/2}, x_i x_i^\top \rangle < 1$, or a consecutive of $p$ iterations fail to improve the minimum eigenvalue of the current solution.[10] Finally, a safeguard is added to record the history of all solutions $\widehat{s}$ appeared throughout the iterations. The algorithm also terminates once the same solution is visited twice.

To find the relaxed continuous solution $\pi$, we use the projected entropic mirror descent Algorithm 3. We use backtracking line search[11] for *differentiable* objectives (e.g., $f_A(\Sigma) = \mathrm{tr}(\Sigma^{-1})/p$ and $g_D(\Sigma) = -1/p \cdot \log \det \Sigma$) and step length $\eta_t = \gamma_0/\sqrt{t+1}$ for *non-differentiable* objectives (e.g., $f_E(\Sigma) = \|\Sigma^{-1}\|_{\mathrm{op}}$), where $\gamma_0$ is chosen so that the algorithm does not overshoot too much. In practice, we start with $\gamma_0 = 0$ and half it (i.e., $\gamma_0 \leftarrow \gamma_0/2$) whenever $f(\omega^{(t+1)}) \geq 2f(\omega^{(t)})$. We stop the algorithm after 100 iterations for differentiable objectives, and after 1000 iterations for non-differentiable objectives.

---
[9] A similar landscape also appears in stochastic gradient methods. For instance, given the ridge regression problem, although interior point or ellipsoid method gives polynomial time algorithm, in practice, one still prefer stochastic gradient methods such as SDCA [40] which depend on the properties of the covariance matrix.

[10] Of course, if the current solution $\widehat{s}$ leads to $\sum_i \widehat{s}_i x_i x_i^\top$ that is not full rank, then we always continue to the next iteration.

[11] In backtracking line search, for every iteration $t$ a preliminary step size of $\eta_t = 1$ is used and the step size is repeatedly halved until the Armijo-Goldstein condition $f(\omega^{(t+1)}) \leq f(\omega^{(t)}) + 0.5 \langle g^{(t)}, \omega^{(t+1)} - \omega^{(t)} \rangle$ is satisfied, where $\omega^{(t+1)}$ is the (projected) next step under step size $\eta_t$.

We compare our proposed algorithm with several previous works listed below.

- *Uniform sampling (*UNIFORM*)*: sample $k$ coordinates from $[n]$ uniformly at random without replacement. The sampling is repeated for 10 times and the best objective in the 10 samples is reported.

- *Weighted sampling (*WEIGHTED*)*: sample $k$ coordinates from $[n]$ without replacement according to the distribution $\pi^*/k$, where $\pi^*$ is the optimal (continuous) solution to Eq. (1.3). The sampling is repeated for 10 times and the best objective in the 10 samples is reported.

- *Fedorov's exchange (*FEDOROV*)*: the Fedorov's exchange algorithm [30, 35] is a popular *heuristic* widely used in statistical computing of optimal designs. The algorithm starts with a random subset of $k$ points and at each iteration selects a pair of points for exchange such that the objective $f$ is minimizes over all such changes. In our experiments we limit the maximum number of changes to 1000, or terminate the algorithm whenever no such exchanges improve the objective.

- *Greedy removal (*GREEDY*)*: the greedy removal procedure starts with the full set $S = [n]$ and removes one coordinate at a time so that the objective is minimized over all such single removals; the algorithm is accurate in most practical applications at the cost of quadratic running time in terms of $n$, making at less practical for large design pools.

    Note that, the greedy method has a provably guarantee for $f_A$ and $f_E$ criteria, but with a large $\frac{n-p+1}{k-p+1}$ factor approximation rate [8]. Its theoretical guarantees for other optimality criteria are unknown.

In the above list, we limit our attention to general-purpose algorithms that can (at least in practice) handle arbitrary optimality crteria, and skip methods that are designed specifically for certain objectives (e.g., submodular optimization for $f_D$ and dual volume sampling [8, 32] for $f_A$ and $f_D$). We also only consider algorithms that can handle "frequentist" objectives which are infinity when $\Sigma = \sum_i \widehat{s}_i x_i x_i^\top$ is singular, thus excluding algorithms like [13, 19] that require the objective $f$ to be well-defined and finite-valued for all positive semi-definite matrices.

## 4.2 Data and Objectives

We synthesize a $n \times p$ design pool $X$ as follows:

$$X = \begin{bmatrix} X_A & 0_{(n/2)\times(p/2)} \\ 0_{(n/2)\times(p/2)} & X_B \end{bmatrix},$$

where $X_A$ is an $(n/2) \times (p/2)$ random Gaussian matrix, re-scaled so that the eigenvalues of $X_A^\top X_A$ satisfy a quadratic decay: $\sigma_j(X_A^\top X_A) \propto j^{-2}$; $X_B$ is an $(n/2) \times (p/2)$ random Gaussian matrix, re-scaled so that the eigenvalues of $X_B^\top X_B$ satisfy a linear decay: $\sigma_j(X_B^\top X_B) \propto j^{-1}$. Such synthetic setting is carefully selected: the decay of the eigenvalues in the Gaussian designs mean that there are important design points $x_i$, and hence uniform sampling may not work well; on the other hand, the split of the two "signal" matrices $X_A$ and $X_B$ demands a careful balance between points allocated in $A$ and $B$, because algorithms that focus solely on one set would produce close to singular designs and thus suffer high objective loss.

Five objectives are selected: the A-optimality $f_A(\Sigma) = \mathrm{tr}(A^{-1})/p$, the D-optimality $f_D(\Sigma) = \det \Sigma^{-1/p}$, the E-optimality $f_E(\Sigma) = \|\Sigma^{-1}\|_{\mathrm{op}}$, the V-optimality $f_V(\Sigma) = \mathrm{tr}(X\Sigma^{-1}X^\top)/n$ and the G-optimality $f_G(\Sigma) = \max \mathrm{diag}(X\Sigma^{-1}X^\top)$.

## 4.3 Results

In Tables 2, 3 and 4 we report performance of our continuous relaxation and the swapping algorithm, together with other competitors mentioned in Section 4.1. We also report the running time (in brackets) of each algorithm, except for the uniform sampling algorithm which finishes instantly on all data sets.

The simulation results suggest that our algorithm (SWAPPING) consistently outperforms uniform sampling (UNIFORM), weighted sampling (WEIGHTED) and Fedorov's exchange algorithm (FEDOROV) for all experimental settings and objectives, especially in cases where $k$ is close to $p$. Recall these are the cases when UNIFORM and WEIGHTED perform very badly due to statistical fluctuation of the sampling procedures.

Our swapping algorithm performs comparable or slightly worse than GREEDY. However, our algorithm is computationally efficient and can handle a wide range of objectives and input sizes. In contrast, the time complexity of the greedy algorithm scales quadratically or even cubically (e.g., the G-optimality) with the number of input points $n$ and soon becomes intractable for intermediate-sized inputs (e.g., $n > 10^4$).

# 5 Concluding Remarks and Open Questions

In this paper we have proposed a general framework for optimizing convex objectives subject to discrete (cardinality) constraints that have wide applications in statistics and machine learning, such as in classical and Bayesian experimental design and active learning. Our algorithm is computationally efficient both in theory and practice, and enjoys rigorous near-optimal performance guarantees.

To conclude this paper, we mention two open directions for future research:

**More efficient methods for continuous optimization.** In Section 3 we showed that the continuous relaxation of the discrete optimization problem is convex, and presented an entropic mirror descent algorithm to solve it. While reducing the computational complexity of such solvers is not the main focus of this paper, it is observed in our simulations that this convex optimization is actually the bottleneck in practice; in contrast, the time complexity of our rounding algorithm is negligible.[12] It is thus an important question to further reduce the computational load of solving the continuous optimization problem in Eq. (1.3). The work of [31] might be helpful, and they considered the dual problem and an approximate interior-point solver for the D-optimality criterion.

**Negative results.** Our main result (Theorem 1.4) shows that in order to achieve $(1+\varepsilon)$-approximation with respect to the discrete solution of Eq. (1.1), the "budget parameter" $k$ needs to be at least $\Omega(p/\varepsilon^2)$.

---

[12]For instance, in Tables 2, 3 and 4, it can be observed that WEIGHTED and SWAPPING have very similar running time. This means the computational overhead (i.e., the rounding process) introduced by the swapping algorithm is negligible.

We believe this bound $\Omega(p/\varepsilon^2)$ is tight for any continuous relaxation based approaches at least for the E-optimality criterion, and thus also tight for the general objectives $f : \mathbb{S}_p^+ \to \mathbb{R}^+$ that satisfy assumptions (A1)-(A3). Such tightness result can be established by the construction of $p$-dimensional vectors $\{x_i\}_{i=1}^n$ such that

- $\frac{k}{n} \sum_{i=1}^n x_i x_i^\top = I_{p \times p}$, but
- no size-$k$ subset $S \subseteq [n]$ satisfy $\sum_{i \in S} x_i x_i^\top \succeq (1 - \varepsilon) I_{p \times p}$ unless $k = \Omega(p/\varepsilon^2)$.

We conjecture that this is possible but have not been able to find a simple proof. Meanwhile, if the second requirement is replaced with its "two-sided" variant, namely $(1-\varepsilon)I_{p\times p} \preceq \sum_{i\in S} x_i x_i^\top \preceq (1+\varepsilon)I_{p\times p}$, then such a construction is possible and proved by [10] using spectral graph theory. It seems very difficult to adapt the proof of [10] to establish our "one-sided" conjecture.

## Appendix A  Missing Proofs

### A.1  Proof of Claim 2.4

**Claim 2.4** (closed form $\ell_{1/2}$ strategy). *Assume without loss of generality that $A_0 = (c_0 I + \alpha Z_0)^{-2}$ for some $c_0 \in \mathbb{R}$ and symmetric matrix $Z_0$ such that $c_0 I + \alpha Z_0 \succ 0$. Then,*

$$A_t = \left( c_t I + \alpha Z_0 + \alpha \sum_{\ell=0}^{t-1} F_\ell \right)^{-2}, \qquad t = 1, 2, \ldots, \tag{2.5}$$

*where $c_t \in \mathbb{R}$ is the unique constant so that $c_t I + \alpha Z_0 + \alpha \sum_{\ell=0}^{t-1} F_\ell \succ 0$ and $\mathrm{tr}(A_t) = 1$.*

*Proof of Claim 2.4.* We first show that for any symmetric matrix $Z \in \mathbb{R}^{p \times p}$, there exists unique $c \in \mathbb{R}$ such that $\alpha Z + cI \succ 0$ and $\mathrm{tr}[(\alpha Z + cI)^{-2}] = 1$. By simple asymptotic analysis, $\lim_{c \to (-\alpha \lambda_{\min}(Z))^+} \mathrm{tr}[(\alpha Z + cI)^{-2}] = +\infty$ and $\lim_{c \to +\infty} \mathrm{tr}[(\alpha Z + cI)^{-2}] = 0$. Because $\mathrm{tr}[(\alpha Z + cI)^{-2}]$ is a continuous and strictly decreasing function in $c$ on the open interval $(-\alpha \lambda_{\min}(Z), +\infty)$, we conclude that there must exist a unique $c$ such that $\mathrm{tr}[(\alpha Z + cI)^{-2}] = 1$. The range of $c$ also ensures that $\alpha Z + cI \succ 0$.

We now use induction to prove this proposition. For $t = 0$ the proposition is obviously correct. We shall now assume that the proposition holds true for $A_{t-1}$ (i.e., $A_{t-1} = (c_{t-1}I + \alpha Z_0 + \sum_{\ell=0}^{t-2} F_\ell)^{-2}$) for some $t \geq 1$, and try to prove the same for $A_t$.

The KKT condition of the optimization problem and the gradients of the Bregman divergence $\Delta_\psi$ yields

$$\nabla \psi(A_t) - \nabla \psi(A_{t-1}) + \alpha F_{t-1} - d_t I = 0, \tag{A.1}$$

where the $d_t I$ term arises from the Lagrangian multiplier and $d_t \in \mathbb{R}$ is the unique number that makes $-\nabla \psi(A_{t-1}) + \alpha F_t - d_t I \preceq 0$ (because $\nabla \psi(A_t) \succeq 0$) and $\mathrm{tr}(A_t) = \mathrm{tr}((\nabla \psi)^{-1}(\nabla \psi(A_{t-1}) + d_t I - \alpha F_{t-1})) = 1$. Re-organizing terms in Eq. (A.1) and invoking the induction hypothesis we have

$$A_t = (\nabla \psi)^{-1} \left( \nabla \psi(A_{t-1}) + d_t I - \alpha F_{t-1} \right)$$
$$= (\nabla \psi)^{-1} \left( -c_{t-1} I - \alpha Z_0 - \alpha \sum_{\ell=0}^{t-1} F_\ell + d_t I \right).$$

Because $d_t$ is the unique number that ensures $A_t \succeq 0$ and $\mathrm{tr}(A_t) = 1$, and $Z_0 + \sum_{\ell=0}^{t-1} F_\ell \succeq 0$, it must hold that $-c_{t+1} + d_t = c_t$. Subsequently, $\nabla \psi(A_t) = -(A_t^{-1/2}) = -c_t I - \alpha Z - \alpha \sum_{\ell=0}^{t-1} F_\ell$. The claim is thus proved by raising both sides of the identity to the power of $-2$. $\square$

## A.2 Proof of Claim 2.10

**Claim 2.10.** *Suppose $P_t^\top A_t^{1/2} P_t = [b\ d; d\ c] \in \mathbb{R}^{2\times 2}$ and $2\alpha \langle A_t^{1/2}, v_t v_t^\top \rangle < 1$. Then*

$$(J + P_t^\top A_t^{1/2} P_t)^{-1} = \left(J + \begin{bmatrix} b & d \\ d & c \end{bmatrix}\right)^{-1} \succeq \left(J + \begin{bmatrix} 2b & 0 \\ 0 & 2c \end{bmatrix}\right)^{-1}.$$

*Proof of Claim 2.10.* Define $R = \begin{bmatrix} b & -d \\ -d & c \end{bmatrix}$. Because $P_t^\top A_t^{1/2} P_t = \begin{bmatrix} b & d \\ d & c \end{bmatrix}$ is positive semi-definite, we conclude that $R$ is also positive semi-definite and hence can be written as $R = QQ^\top$. To prove Claim 2.10, we only need to establish the positive semi-definiteness of the following difference matrix:

$$\left(J + \begin{bmatrix} b & d \\ d & c \end{bmatrix}\right)^{-1} - \left(J + \begin{bmatrix} 2b & 0 \\ 0 & 2c \end{bmatrix}\right)^{-1}$$

$$= \left(J + \begin{bmatrix} 2b & 0 \\ 0 & 2c \end{bmatrix} - \begin{bmatrix} b & -d \\ -d & c \end{bmatrix}\right)^{-1} - \left(J + \begin{bmatrix} 2b & 0 \\ 0 & 2c \end{bmatrix}\right)^{-1}$$

$$= \left(J + \begin{bmatrix} 2b & 0 \\ 0 & 2c \end{bmatrix}\right)^{-1} Q \left(I - Q^\top \left(J + \begin{bmatrix} 2b & 0 \\ 0 & 2c \end{bmatrix}\right)^{-1} Q\right)^{-1} Q^\top \left(J + \begin{bmatrix} 2b & 0 \\ 0 & 2c \end{bmatrix}\right)^{-1}.$$

Here in the last equality we again use the Woodbury matrix identity. It is clear that to prove the positive semi-definiteness right-hand side of the above equality, It suffices to show $Q^\top (J + \mathrm{diag}(2b, 2c))^{-1} Q \prec I$. By standard matrix analysis and the fact that $J = \mathrm{diag}(1, -1)$,

$$Q^\top \left(J + \begin{bmatrix} 2b & 0 \\ 0 & 2c \end{bmatrix}\right)^{-1} Q = Q^\top \begin{bmatrix} (1+2b)^{-1} & 0 \\ 0 & -(1-2c)^{-1} \end{bmatrix} Q$$

$$\stackrel{(a)}{\preceq} Q^\top \begin{bmatrix} (1+2b)^{-1} & 0 \\ 0 & 0 \end{bmatrix} Q \preceq \frac{\|QQ^\top\|_{\mathrm{op}}}{1+2b} \cdot I$$

$$\stackrel{(b)}{\preceq} \frac{\max\{2b, 2c\}}{1+2b} \cdot I \stackrel{(c)}{\prec} I.$$

Some steps in the above derivation require additional explanation. In (a), we use the fact that $2c = 2\alpha \langle A_t^{1/2}, v_t v_t^\top \rangle < 1$, and hence $(1-2c)^{-1} > 0$; in (b), we use the fact that $QQ^\top = \begin{bmatrix} b & -d \\ -d & c \end{bmatrix} \preceq \begin{bmatrix} 2b & 0 \\ 0 & 2c \end{bmatrix}$; finally, (c) holds because $b = 2\alpha \langle A_t^{1/2}, u_t u_t^\top \rangle \geq 0$, $\frac{2b}{1+2b} < 1$ and $2c < 1$. The proof of Claim 2.10 is thus completed. $\square$

## A.3 Proof of Claim 2.11

**Claim 2.11.** *Suppose $Z \succeq 0$ is a $p$-dimensional PSD matrix with $\lambda_{\min}(Z) \leq 1$. Let $A = (\alpha Z + cI)^{-2}$, where $c \in \mathbb{R}$ is the unique real number such that $A \succeq 0$ and $\mathrm{tr}(A) = 1$. Then*

1. $\alpha \langle A^{1/2}, Z \rangle \leq p + \alpha \sqrt{p}$;

2. $\langle A, Z \rangle \leq \sqrt{p}/\alpha + \lambda_{\min}(Z)$.

*Proof of Claim 2.11.* For any orthogonal matrix $U$, the transform $Z \mapsto UZU^\top$ leads to $X \mapsto UXU^\top$ and $X^{1/2} \mapsto UX^{1/2}U^\top$; thus both inner products are invariant to orthogonal transform of $Z$. Therefore, we may assume without loss of generality that $Z = \mathrm{diag}(\sigma_1, \ldots, \sigma_p)$ for $\lambda_1 \geq \cdots \lambda_p \geq 0$, because $Z \succeq 0$. Subsequently,

$$\alpha \langle A^{1/2}, Z \rangle = \sum_{i=1}^p \frac{\alpha \lambda_i}{\alpha \lambda_i + c} = p - c \cdot \sum_{i=1}^p \frac{1}{\alpha \lambda_i + c}.$$

If $c \geq 0$, then $\alpha \langle A^{1/2}, Z \rangle \leq p$ and the first property is clearly true. For the case of $c < 0$, note that $c$ must be strictly larger than $-\alpha \lambda_p$, as we established in Claim 2.4. Subsequently, by the Cauchy-Schwarz inequality,

$$\alpha \langle A^{1/2}, Z \rangle = p - c \cdot \sum_{i=1}^p \frac{1}{\alpha \lambda_i + c} \leq p - c \cdot \sqrt{p} \cdot \sqrt{\sum_{i=1}^p \frac{1}{(\alpha \lambda_i + c)^2}}.$$

Because $\lambda_p = \lambda_{\min}(Z) \leq 1$ and $\mathrm{tr}(A) = \mathrm{tr}[(\alpha Z + cI)^{-2}] = 1$, we have that $c \geq -\alpha$ and $\sqrt{\sum_{i=1}^p (\alpha \lambda_i + c)^{-2}} = 1$. Therefore, $\alpha \langle A^{1/2}, Z \rangle \leq p + \alpha \sqrt{p}$, which establishes the first property in Claim 2.11.

We next turn to the second property. Using similar analysis, we have

$$\alpha \langle Z, A \rangle = \sum_{i=1}^p \frac{\alpha \lambda_i}{(\alpha \lambda_i + c)^2} = \sum_{i=1}^p \frac{1}{\alpha \lambda_i + c} - c \cdot \sum_{i=1}^p \frac{1}{(\alpha \lambda_i + c)^2}$$

$$\leq \sqrt{p} \cdot \sqrt{\sum_{i=1}^p \frac{1}{(\alpha \lambda_i + c)^2}} - c \cdot \sum_{i=1}^p \frac{1}{(\alpha \lambda_i + c)^2} \leq \sqrt{p} - c.$$

Property 2 is then proved by noting that $c > -\lambda_{\min}(Z)$. □

### A.4 Proof of correctness of Algorithm 4

Recall the KL divergence function $\mathrm{KL}(y \| \omega) := \sum_i y_i \log \frac{y_i}{\omega_i}$.

**Claim A.1.** *The output of Algorithm 4 is exactly the projection $\omega' = \arg\min_{y \in \Delta_n, y_i \leq b} \mathrm{KL}(y \| \omega)$ in KL divergence, provided that the input $\omega$ itself is in the probabilistic simplex $\Delta_n$.*

*Proof.* We first show that if $\omega_i > \omega_j$ then $\omega'_i \geq \omega'_j$. Assume the contrary that $\omega_i > \omega_j$ and $\omega'_i < \omega'_j$. Define $\omega'' = \omega'$ except that the $i$th and $j$th components are swapped; that is, $\omega''_i = \omega'_j$ and $\omega''_j = \omega'_i$. It is clear that $\omega'' \in \Delta_n$ and satisfies the box constraint $\|\omega''\|_\infty \leq b$. In addition,

$$\mathrm{KL}(\omega'' \| \omega) - \mathrm{KL}(\omega' \| \omega) = \omega'_i \log \frac{\omega_i}{\omega_j} + \omega'_j \log \frac{\omega_j}{\omega_i} = (\omega'_i - \omega'_j) \log \frac{\omega_i}{\omega_j} < 0,$$

which violates the optimality of $\omega'$. □

We next consider the Lagrangian multiplier of the constrained optimization problem

$$\mathcal{L}(y; \mu, \lambda) = \sum_{i=1}^n y_i \log \frac{y_i}{\omega_i} + \mu \left( \sum_{i=1}^n y_i - 1 \right) + \sum_{i=1}^n \lambda_i (y_i - b).$$

By KKT condition, we have that $\partial \mathcal{L}(\omega')/\partial y_i = \log(y_i/\omega_i) + (1 + \mu + \lambda_i) = 0$. By complementary slackness, if $\omega'_i < b$ then $\lambda_i = 0$ and hence there exists a unique $C > 0$ such that $\omega'_i = C \cdot \omega_i$ holds for all $\omega'_i < b$. Combining this fact with the monotonic correspondence between $\omega'$ and $\omega$, one only needs to search for the exact number of components in $\omega'$ that are equal to $b$, and compute the unique constant $C$ and the remaining coordinates. Since there are at most $n$ such possibilities, by a linear scan over the choices we can choose the solution that gives rise to the minimum KL divergence. This is exactly what is computed in Algorithm 4: we have a $O(n)$ time linear scan over parameter $q$ (meaning that there are exactly $q - 1$ components that are equal to $b$, proceeded by a $O(n \log n)$-time pre-processing to sort the coordinates of $\omega$.

## Acknowledgements

We thank Adams Wei Yu for helpful discussions regarding the implementation of the entropic mirror descent solver for the continuous (convex) relaxation problem, thank Aleksandar Nikolov, Shayan Oveis Gharan, and Mohit Singh for discussions on the references. This work is supported by NSF CCF-1563918, NSF CAREER IIS-1252412 and AFRL FA87501720212.

Table 2: Results on synthetic data with $n = 1000$ and $p = 50$. Numbers in brackets indicate the running time (in seconds) for each algorithm (omitted for UNIFORM, which does not take significant running time). Our proposed algorithm (Alg. 1) appears as SWAPPING. The running time for both WEIGHTED and SWAPPING takes into account the time for continuous optimization.

|  | $f_A$ |  | $f_D$ |  | $f_E$ |  | $f_V$ |  | $f_G$ |  |
|---|---|---|---|---|---|---|---|---|---|---|
| $k = 1.2p = 60$ |  |  |  |  |  |  |  |  |  |  |
| UNIFORM | 76.0 |  | 11.6 |  | 776 |  | 303 |  | 1654 |  |
| WEIGHTED | Inf | *(1.8)* | 7.42 | *(0.5)* | Inf | *(4.7)* | 177 | *(5.1)* | 982 | *(9.2)* |
| FEDOROV | Inf | *(79)* | Inf | *(0.0)* | 700 | *(17×10³)* | Inf | *(0.0)* | 202 | *(4.7×10³)* |
| GREEDY | 16.3 | *(11)* | 5.91 | *(11)* | 73.4 | *(0.6×10³)* | 60.3 | *(12)* | 152 | *(1.1×10³)* |
| SWAPPING | 19.1 | *(5.1)* | 6.04 | *(3.1)* | 114 | *(7.3)* | 60.7 | *(7.5)* | 142 | *(12)* |
| $k = 1.5p = 75$ |  |  |  |  |  |  |  |  |  |  |
| UNIFORM | 40.5 |  | 9.81 |  | 411 |  | 131 |  | 398 |  |
| WEIGHTED | 25.9 | *(1.2)* | 6.46 | *(0.3)* | Inf | *(3.8)* | 82.6 | *(3.7)* | 246 | *(11)* |
| FEDOROV | 15.3 | *(0.1×10³)* | 5.95 | *(0.1×10³)* | 285 | *(38×10³)* | 56.5 | *(0.1×10³)* | 136 | *(6.7×10³)* |
| GREEDY | 13.1 | *(11)* | 5.47 | *(11)* | 49.5 | *(0.6×10³)* | 49.3 | *(12)* | 104 | *(1.1×10³)* |
| SWAPPING | 15.2 | *(4.2)* | 5.62 | *(3.4)* | 61.5 | *(6.9)* | 49.4 | *(6.9)* | 97.5 | *(14)* |
| $k = 2p = 100$ |  |  |  |  |  |  |  |  |  |  |
| UNIFORM | 27.2 |  | 8.40 |  | 202 |  | 92.7 |  | 246 |  |
| WEIGHTED | 14.9 | *(1.7)* | 5.66 | *(0.5)* | Inf | *(5.0)* | 56.4 | *(4.9)* | 156 | *(11)* |
| FEDOROV | 13.8 | *(0.1×10³)* | 5.70 | *(0.1×10³)* | 148 | *(0.9×10³)* | 47.8 | *(0.2×10³)* | 114 | *(11×10³)* |
| GREEDY | 11.7 | *(11)* | 5.19 | *(11)* | 38.3 | *(0.6×10³)* | 43.3 | *(12)* | 81.9 | *(1.1×10³)* |
| SWAPPING | 13.1 | *(7.4)* | 5.21 | *(6.1)* | 47.8 | *(9.1)* | 44.5 | *(9.4)* | 77.9 | *(16)* |
| $k = 3p = 150$ |  |  |  |  |  |  |  |  |  |  |
| UNIFORM | 23.3 |  | 7.55 |  | 155 |  | 70.9 |  | 168 |  |
| WEIGHTED | 12.2 | *(2.0)* | 5.24 | *(0.5)* | 266 | *(4.6)* | 46.0 | *(6.3)* | 96.0 | *(12)* |
| FEDOROV | 12.3 | *(0.3×10³)* | 5.57 | *(0.3×10³)* | 135 | *(0.3×10³)* | 44.8 | *(0.3×10³)* | 87.4 | *(25×10³)* |
| GREEDY | 10.9 | *(10)* | 5.08 | *(10)* | 33.9 | *(0.6×10³)* | 40.5 | *(11)* | 70.8 | *(1.1×10³)* |
| SWAPPING | 11.6 | *(8.4)* | 5.08 | *(9.5)* | 40.3 | *(10.9)* | 40.4 | *(13.4)* | 66.6 | *(18.4)* |
| $k = 5p = 250$ |  |  |  |  |  |  |  |  |  |  |
| UNIFORM | 18.7 |  | 7.23 |  | 122 |  | 58.0 |  | 134 |  |
| WEIGHTED | 11.5 | *(1.5)* | 5.23 | *(0.41)* | 115 | *(5.6)* | 41.6 | *(3.0)* | 90.2 | *(9.3)* |
| FEDOROV | 23.1 | *(0.8)* | 5.70 | *(0.5×10³)* | 93.6 | *(1.1×10³)* | 59.1 | *(0.5)* | 89.4 | *(26×10³)* |
| GREEDY | 11.0 | *(11)* | 5.17 | *(11)* | 34.5 | *(0.6×10³)* | 39.7 | *(11)* | 66.9 | *(1.1×10³)* |
| SWAPPING | 11.4 | *(11)* | 5.17 | *(15)* | 36.2 | *(17)* | 39.8 | *(16)* | 66.7 | *(23)* |

Table 3: Results on synthetic data with $n = 5000$ and $p = 50$. Numbers in brackets indicate the running time (in seconds) for each algorithm (omitted for UNIFORM, which does not take significant running time). Our proposed algorithm (Alg. 1) appears as SWAPPING. The running time for both WEIGHTED and SWAPPING takes into account the time for continuous optimization.

| | $f_A$ | | $f_D$ | | $f_E$ | | $f_V$ | | $f_G$ | |
|---|---|---|---|---|---|---|---|---|---|---|
| $k = 1.2p = 60$ | | | | | | | | | | |
| UNIFORM | 32.1 | | 3.89 | | 40.0 | | 53.2 | | 300 | |
| WEIGHTED | 1.83 | *(3.3)* | 1.04 | *(2.6)* | 13.5 | *(29)* | 4.13 | *(13)* | 203 | *(80)* |
| FEDOROV | 9.68 | *(0.0)* | 6.86 | *(0.0)* | Inf | *(0.0)* | 94.4 | *(0.0)* | 26.1 | *(1.4×10³)* |
| GREEDY | 1.27 | *(9.5)* | 0.88 | *(9.6)* | 4.46 | *(0.6×10³)* | 3.24 | *(10)* | 17.9 | *(5.5×10³)* |
| SWAPPING | 1.31 | *(3.5)* | 0.87 | *(2.8)* | 3.41 | *(29)* | 3.11 | *(13)* | 19.6 | *(80)* |
| $k = 1.5p = 75$ | | | | | | | | | | |
| UNIFORM | 8.05 | | 3.91 | | 109 | | 17.9 | | 155 | |
| WEIGHTED | 1.29 | *(3.5)* | 0.96 | *(2.9)* | 6.93 | *(30)* | 3.48 | *(19)* | 51.0 | *(77)* |
| FEDOROV | 25.1 | *(0.0)* | Inf | *(0.0)* | 41.7 | *(1.0×10³)* | 3.2 | *(0.0)* | 25.0 | *(0.6×10³)* |
| GREEDY | 1.20 | *(9.7)* | 0.87 | *(9.8)* | 4.15 | *(0.6×10³)* | 2.97 | *(10)* | 16.8 | *(5.4×10³)* |
| SWAPPING | 1.30 | *(3.8)* | 0.93 | *(3.1)* | 2.39 | *(30)* | 3.07 | *(19)* | 15.1 | *(78)* |
| $k = 2p = 100$ | | | | | | | | | | |
| UNIFORM | 6.34 | | 4.17 | | 18.0 | | 15.4 | | 127 | |
| WEIGHTED | 1.20 | *(3.5)* | 0.89 | *(2.6)* | 4.80 | *(28)* | 2.96 | *(17)* | 28.5 | *(82)* |
| FEDOROV | 1.33 | *(1.9)* | 0.92 | *(2.6)* | 16.7 | *(1.5×10³)* | 3.06 | *(2.4)* | 17.1 | *(1.4×10³)* |
| GREEDY | 1.16 | *(9.3)* | 0.86 | *(9.6)* | 3.32 | *(0.6×10³)* | 2.85 | *(10)* | 13.0 | *(5.4×10³)* |
| SWAPPING | 1.22 | *(3.8)* | 0.86 | *(3.0)* | 2.10 | *(28)* | 2.91 | *(17)* | 12.9 | *(83)* |
| $k = 3p = 150$ | | | | | | | | | | |
| UNIFORM | 5.03 | | 3.41 | | 22.3 | | 12.9 | | 105 | |
| WEIGHTED | 1.22 | *(3.0)* | 0.90 | *(2.9)* | 4.12 | *(30)* | 2.99 | *(8.1)* | 18.1 | *(82)* |
| FEDOROV | 1.20 | *(6.2)* | 0.97 | *(7.0)* | 12.9 | *(5.0×10³)* | 3.25 | *(4.8)* | 18.7 | *(2.1×10³)* |
| GREEDY | 1.15 | *(9.4)* | 0.88 | *(9.6)* | 3.08 | *(0.6×10³)* | 2.87 | *(10)* | 12.4 | *(5.2×10³)* |
| SWAPPING | 1.18 | *(3.6)* | 0.89 | *(3.5)* | 1.92 | *(30)* | 2.87 | *(8.6)* | 11.9 | *(82)* |
| $k = 5p = 250$ | | | | | | | | | | |
| UNIFORM | 4.76 | | 3.30 | | 11.6 | | 10.9 | | 67.7 | |
| WEIGHTED | 1.24 | *(3.4)* | 0.93 | *(3.1)* | 3.43 | *(33)* | 2.99 | *(6.5)* | 20.0 | *(82)* |
| FEDOROV | 5.99 | *(0.0)* | 3.11 | *(0.0)* | 11.5 | *(5.6×10³)* | 3.52 | *(8.8)* | 18.6 | *(5.6×10³)* |
| GREEDY | 1.21 | *(9.7)* | 0.92 | *(9.9)* | 2.93 | *(0.6×10³)* | 2.97 | *(10.5)* | 12.6 | *(5.5×10³)* |
| SWAPPING | 1.27 | *(4.5)* | 0.92 | *(4.2)* | 1.96 | *(34)* | 2.98 | *(7.4)* | 12.1 | *(83)* |

Table 4: Results on synthetic data with $n = 10000$ and $p = 50$. Numbers in brackets indicate the running time (in seconds) for each algorithm (omitted for UNIFORM, which does not take significant running time). Our proposed algorithm (Alg. 1) appears as SWAPPING. The running time for both WEIGHTED and SWAPPING takes into account the time for continuous optimization. All results of the Fedorov exchange algorithm and G-optimality results for the greedy algorithm are omitted because the algorithms took too long to converge.

|  | $f_A$ |  | $f_D$ |  | $f_E$ |  | $f_V$ |  | $f_G$ |  |
|---|---|---|---|---|---|---|---|---|---|---|
| $k = 1.2p = 60$ |  |  |  |  |  |  |  |  |  |  |
| UNIFORM | 69.7 |  | 11.7 |  | 687 |  | 339 |  | 1818 |  |
| WEIGHTED | Inf | $(0.1 \times 10^3)$ | 5.93 | $(30)$ | Inf | $(0.3 \times 10^3)$ | 132 | $(0.4 \times 10^3)$ | 1652 | $(0.6 \times 10^3)$ |
| GREEDY | 11.6 | $(1.1 \times 10^3)$ | 4.61 | $(1.1 \times 10^3)$ | 60.3 | $(6.6 \times 10^4)$ | 47.3 | $(1.2 \times 10^3)$ | - | - |
| SWAPPING | 13.7 | $(0.1 \times 10^3)$ | 5.10 | $(37)$ | 60.9 | $(0.3 \times 10^3)$ | 50.3 | $(0.4 \times 10^3)$ | 152 | $(0.6 \times 10^3)$ |
| $k = 1.5p = 75$ |  |  |  |  |  |  |  |  |  |  |
| UNIFORM | 41.0 |  | 10.0 |  | 414 |  | 157 |  | 606 |  |
| WEIGHTED | 15.7 | $(0.2 \times 10^3)$ | 4.87 | $(20)$ | Inf | $(0.3 \times 10^3)$ | 69.1 | $(0.4 \times 10^3)$ | 378 | $(0.5 \times 10^3)$ |
| GREEDY | 9.55 | $(1.1 \times 10^3)$ | 4.21 | $(1.1 \times 10^3)$ | 35.4 | $(6.7 \times 10^4)$ | 39.5 | $(1.2 \times 10^3)$ | - | - |
| SWAPPING | 10.7 | $(0.2 \times 10^3)$ | 4.26 | $(27)$ | 42.4 | $(0.3 \times 10^3)$ | 40.4 | $(0.4 \times 10^3)$ | 95.0 | $(0.5 \times 10^3)$ |
| $k = 2p = 100$ |  |  |  |  |  |  |  |  |  |  |
| UNIFORM | 27.9 |  | 8.41 |  | 165 |  | 97.6 |  | 378 |  |
| WEIGHTED | 10.2 | $(0.1 \times 10^3)$ | 4.43 | $(18)$ | Inf | $(0.2 \times 10^3)$ | 47.5 | $(0.6 \times 10^3)$ | 210 | $(0.8 \times 10^3)$ |
| GREEDY | 8.46 | $(1.1 \times 10^3)$ | 3.99 | $(1.1 \times 10^3)$ | 26.5 | $(6.6 \times 10^4)$ | 34.7 | $(1.2 \times 10^3)$ | - | - |
| SWAPPING | 9.48 | $(0.1 \times 10^3)$ | 4.03 | $(29)$ | 29.9 | $(0.2 \times 10^3)$ | 34.3 | $(0.6 \times 10^3)$ | 73.6 | $(0.8 \times 10^3)$ |
| $k = 3p = 150$ |  |  |  |  |  |  |  |  |  |  |
| UNIFORM | 21.3 |  | 7.23 |  | 172 |  | 74.6 |  | 245 |  |
| WEIGHTED | 9.13 | $(0.2 \times 10^3)$ | 4.03 | $(35)$ | 342 | $(0.4 \times 10^3)$ | 37.0 | $(0.3 \times 10^3)$ | 114 | $(0.5 \times 10^3)$ |
| GREEDY | 7.81 | $(1.1 \times 10^3)$ | 3.88 | $(1.1 \times 10^3)$ | 21.9 | $(6.6 \times 10^4)$ | 32.1 | $(1.2 \times 10^3)$ | - | - |
| SWAPPING | 8.41 | $(0.2 \times 10^3)$ | 3.89 | $(53)$ | 23.3 | $(0.4 \times 10^3)$ | 32.2 | $(0.3 \times 10^3)$ | 62.6 | $(0.5 \times 10^3)$ |